\def\paperID{4160}
\def\confName{CVPR}
\def\confYear{2025}

\def\authorBlock{
Chung-Ho Wu$^1$
\quad
Yang-Jung Chen$^1$
\quad
Ying-Huan Chen$^1$
\quad
Jie-Ying Lee$^1$
\\
Bo-Hsu Ke$^1$
\quad
Chun-Wei Tuan Mu$^1$
\quad
Yi-Chuan Huang$^1$
\quad
Chin-Yang Lin$^1$
\\
Min-Hung Chen$^2$
\quad
Yen-Yu Lin$^1$
\quad
Yu-Lun Liu$^1$\vspace{0.5em}
\\
\centerline{$^1$National Yang Ming Chiao Tung University \quad $^2$NVIDIA}\vspace{0.5em}
\\
{\url{https://kkennethwu.github.io/aurafusion360/}}
}

\newif\ifreview 
\newif\ifarxiv \newcommand{\arxiv}{\arxivtrue}
\newif\ifcamera 
\newif\ifrebuttal 

\arxiv

\pdfoutput=1
\documentclass[10pt,twocolumn,letterpaper]{article}
\ifreview \usepackage[review]{cvpr} \fi
\ifarxiv \usepackage[pagenumbers]{cvpr} \fi
\ifrebuttal \usepackage[rebuttal]{cvpr} \fi
\ifcamera \usepackage{cvpr} \fi

\usepackage{graphicx}	
\usepackage{amsmath}	
\usepackage{amssymb}	
\usepackage{booktabs}
\usepackage{times}
\usepackage{microtype}
\usepackage{epsfig}
\usepackage{caption}
\usepackage{float}
\usepackage{placeins}
\usepackage{color, colortbl}
\usepackage{stfloats}
\usepackage{enumitem}
\usepackage{tabularx}
\usepackage{xstring}
\usepackage{multirow}
\usepackage{xspace}
\usepackage{url}
\usepackage{subcaption}
\usepackage{xcolor}
\usepackage[hang,flushmargin]{footmisc}

\ifcamera \usepackage[accsupp]{axessibility} \fi

% ...

\ifarxiv  \fi

\newcommand{\R}[1]{{%
    \textbf{%
        \ifstrequal{#1}{1}{\textcolor{red}{R#1}}{%
        \ifstrequal{#1}{2}{\textcolor{blue}{R#1}}{%
        \ifstrequal{#1}{3}{\textcolor{magenta}{R#1}}{%
        \ifstrequal{#1}{4}{\textcolor{teal}{R#1}}{%
                           \textcolor{cyan}{R#1}%
        }}}}%
    }%
}}

\definecolor{cvprblue}{rgb}{0.5, 0.5, 1}  

\usepackage{xr-hyper}

\makeatletter
\newcommand*{\addFileDependency}[1]{
  \typeout{(#1)}
  \@addtofilelist{#1}
  \IfFileExists{#1}{}{\typeout{No file #1.}}
}

\makeatother
\newcommand*{\myexternaldocument}[1]{
    \externaldocument{#1}
    \addFileDependency{#1.tex}
    \addFileDependency{#1.aux}
}

\definecolor{cvprblue}{rgb}{0.21,0.49,0.74}
\usepackage[pagebackref,breaklinks,colorlinks,allcolors=cvprblue]{hyperref}
\usepackage[capitalize]{cleveref}
\crefname{section}{Sec.}{Secs.}
\crefname{table}{Table}{Tables}
\crefname{figure}{Fig.}{Figs.}

\ifarxiv \crefname{appendix}{App.}{Apps.}
\else \crefname{appendix}{Suppl.}{Suppls.} \fi

\unless\ifarxiv \myexternaldocument{_supplementary} \fi

\begin{document}
\title{AuraFusion360: Augmented Unseen Region Alignment for Reference-based 360° Unbounded Scene Inpainting}
\author{\authorBlock}

\twocolumn[{%
\renewcommand\twocolumn[1][]{#1}%
\maketitle
\begin{center}
\centering
\captionsetup{type=figure}
\vspace{-6mm}
\resizebox{1.0\textwidth}{!} 
{
\includegraphics[width=\textwidth]{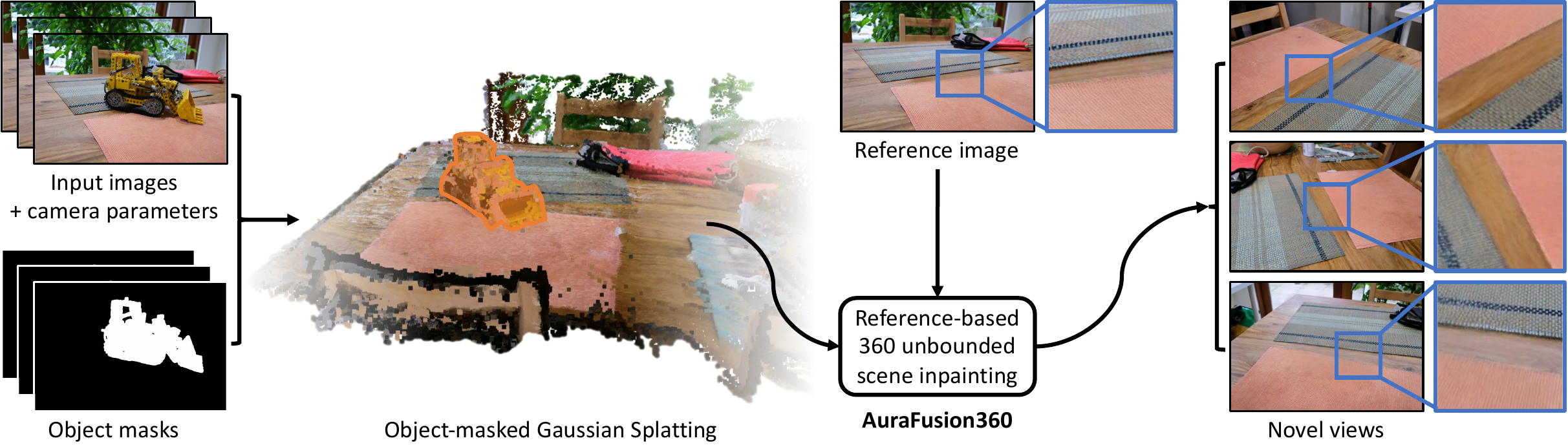}
}
\vspace{-6mm}
\caption{
\textbf{Overview of our reference-based 360° unbounded scene inpainting method.} Given input images with camera parameters, object masks, and a reference image, our AuraFusion360 approach generates an object-masked Gaussian Splatting representation. This representation can then render novel views of the inpainted scene, effectively removing the masked objects while maintaining consistency with the reference image.}
\label{fig:teaser}
\end{center}
}]

\maketitle

\begin{abstract}

Three-dimensional scene inpainting is crucial for applications from virtual reality to architectural visualization, yet existing methods struggle with view consistency and geometric accuracy in 360° unbounded scenes. We present AuraFusion360, a novel reference-based method that enables high-quality object removal and hole filling in 3D scenes represented by Gaussian Splatting. Our approach introduces (1) depth-aware unseen mask generation for accurate occlusion identification, (2) Adaptive Guided Depth Diffusion, a zero-shot method for accurate initial point placement without requiring additional training, and (3) SDEdit-based detail enhancement for multi-view coherence. We also introduce 360-USID, the first comprehensive dataset for 360° unbounded scene inpainting with ground truth. Extensive experiments demonstrate that AuraFusion360 significantly outperforms existing methods, achieving superior perceptual quality while maintaining geometric accuracy across dramatic viewpoint changes.
\end{abstract}

\section{Introduction}
\label{sec:intro}
Three-dimensional scene reconstruction, driven by Neural Radiance Fields~\cite{mildenhall2020nerf} and 3D Gaussian Splatting~\cite{kerbl20233d}, is vital for VR/AR, robotics, and autonomous driving. A key challenge is realistic object removal and hole filling, which is essential for augmented reality and real estate visualization. Inpainting 360° unbounded scenes remains difficult due to the need for multi-view consistency, plausible unseen region extrapolation, and geometric coherence across views.

\cref{fig:teaser} shows our reference-based 360° unbounded scene inpainting approach. Given input images with camera parameters, object masks, and a reference image, our method generates an inpainted 3D scene using Gaussian Splatting~\citep{kerbl20233d,huang20242d} for novel view rendering. We exploit multi-view information and generative models to fill unseen areas, ensuring coherent and plausible results across views. Integrating Gaussian Splatting’s explicit representation with 2D generative inpainting, our method maintains multi-view consistency and geometric accuracy under significant viewpoint changes.

\begin{figure}[t]
    \centering
    \includegraphics[width=1\linewidth]{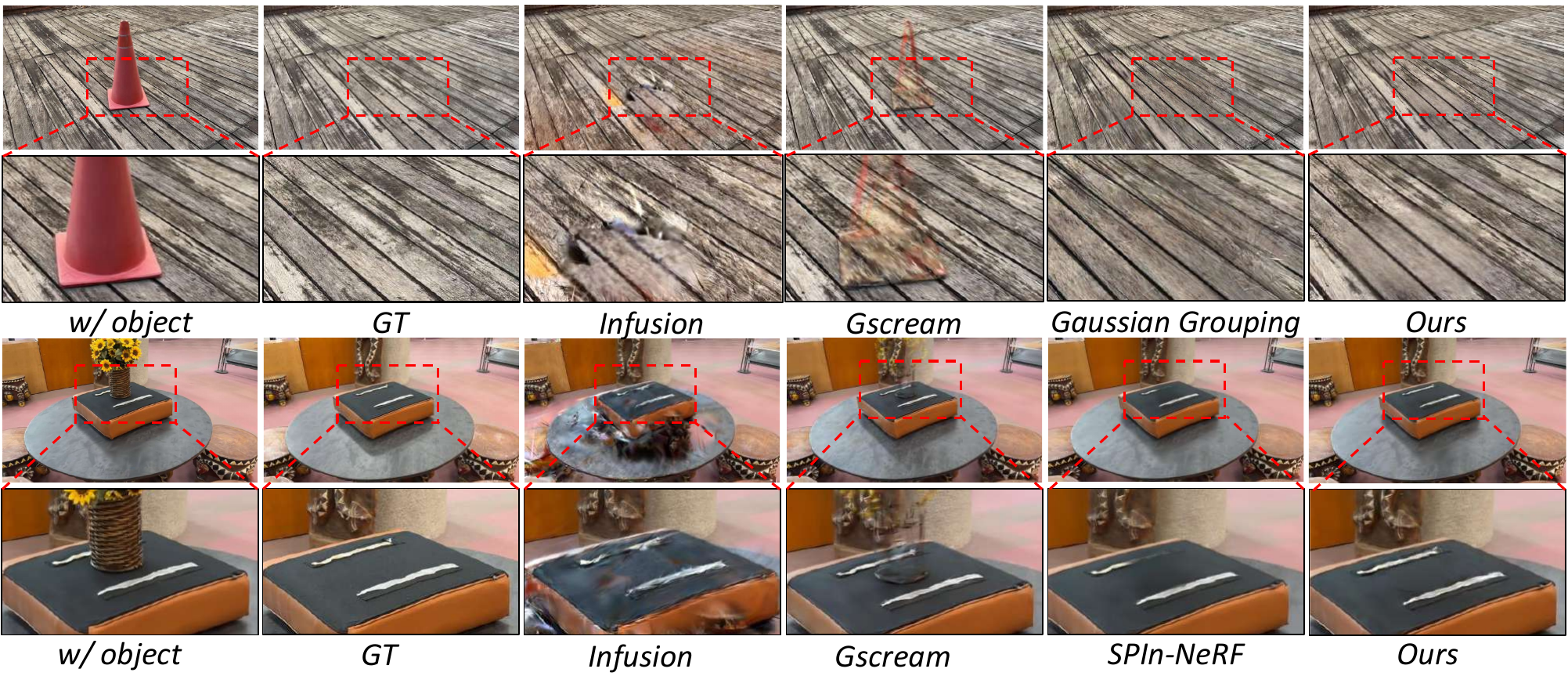}
    \vspace{-6mm}
    \caption{
    \textbf{Comparison with different 3D inpainting approaches.} Existing methods such as SPin-NeRF~\cite{spinnerf} and GScream~\cite{wang2024gscream}, designed for forward-facing scenes, perform poorly in 360° scenarios. Reference-based methods like Infusion~\cite{liu2024infusion} struggle with accurate depth projection, causing fine-tuning artifacts. Gaussian Grouping~\cite{ye2023gaussian} frequently misidentifies unseen regions, reducing inpainting quality. Our AuraFusion360 achieves precise unseen masks and improved depth alignment via Adaptive Guided Depth Diffusion, employing SDEdit~\cite{meng2022sdedit} for diffusion-guided, multi-view consistent RGB generation.
    }
    \vspace{-1mm}
    \label{fig:motivation}
\end{figure}

Several critical challenges in 360° unbounded scene inpainting motivated our approach (\cref{fig:motivation}). Existing methods~\cite{spinnerf, wang2024gscream, mirzaei2023reference, mirzaei2024reffusionreferenceadapteddiffusion}, effective for forward-facing scenes, struggle with extreme viewpoint changes in 360° scenes, resulting in inconsistencies and artifacts. Recent approaches like Gaussian Grouping~\citep{ye2023gaussian} effectively propagate semantic information for object removal, but their reliance on a text-based tracker~\cite{cheng2023tracking} often causes misidentified unseen regions, leading to inaccurate reconstructions.

To address these challenges, we propose a unified pipeline for 360° unbounded scene inpainting using Gaussian Splatting for object removal, depth-aware unseen region detection, and multi-view consistent inpainting. Inspired by Gaussian Grouping~\cite{ye2023gaussian}, our method integrates object-masked attributes into Gaussians for precise removal and reconstructs unseen regions before applying reference-guided inpainting. Unlike methods that directly apply inpainters, causing inconsistencies, we develop Adaptive Guided Depth Diffusion (AGDD) to unproject aligned points from the reference view into unseen regions. These points (1) initialize Gaussians and (2) guide inpainted RGB generation via SDEdit~\cite{meng2022sdedit}, ensuring coherent, high-quality 360° scene restoration.

Integrating these improvements, our framework achieves enhanced geometric accuracy and realism in 360° unbounded scenes. To advance 3D inpainting, we propose a method that improves consistency and provides a benchmark for future research. Our contributions include: 
\begin{itemize} 
\item A depth-aware method leveraging multi-view information to accurately generate unseen masks for 360° unbounded scene inpainting. 
\item Integration of reference view unprojection with SDEdit to produce consistent RGB guidance across views. 
\item A comprehensive framework with a new 360° dataset and capture protocol, supporting high-quality novel view synthesis and quantitative evaluation. 
\end{itemize}


\section{Related Work}
\label{sec:related}

\noindent {\bf NeRF.}
Neural Radiance Fields (NeRF)\citep{mildenhall2020nerf} revolutionized novel view synthesis via differentiable volume rendering\citep{tulsiani2017mvsupervision, henzler2019platonicgan} and positional encoding~\citep{vaswani2017attentionisallyouneed, gehring2017convolutional}. NeRF models improved in efficiency~\citep{liu2020neural, Garbin_2021_ICCV, chen2024improving}, rendering quality~\citep{mipnerf, zhang2020nerf++,meuleman2023progressively}, handling dynamic scenes~\cite{liu2023robust}, and data efficiency~\citep{pixel.nerf, ibrnet, lin2024frugalnerf,su2024boostmvsnerfs}. Despite excelling at view synthesis, NeRF’s implicit representation complicates scene editing. Recent work on object manipulation~\citep{yang2021learning}, stylization~\citep{wang2023nerf, haque2023instruct}, and inpainting~\citep{nerf.in, spinnerf, mirzaei2023reference} struggles with 3D consistency and structural priors, especially in unbounded scenes.

\vspace{3pt}
\noindent {\bf 3D Gaussian Splatting.}
3D Gaussian Splatting (3DGS)~\cite{kerbl20233d} efficiently represents scenes with explicit 3D Gaussians, enabling faster rendering, easier training, and flexible editing\citep{chen2024gaussianeditor}. Recent extensions like Scaffold-GS~\citep{scaffoldgs} enhance efficiency with dynamic anchors, while 2DGS~\citep{huang20242d} refines multi-view geometry. 3DGS has also expanded to dynamic scenes~\citep{yang2024deformable, luiten2023dynamic, Wu_2024_CVPR,fan2025spectromotion} and semantic representations~\citep{ye2023gaussian, qin2023langsplat}, supporting advanced editing and novel view synthesis~\citep{qiu-2024-featuresplatting, huang20242d}. Gaussian-based methods thus offer strong potential for explicit 3D inpainting.
\vspace{3pt}
\noindent {\bf Traditional and learning-based image inpainting.}
Early image inpainting techniques, including PDE-based~\citep{bertalmio2000image}, exemplar-based~\citep{criminisi2004region}, and PatchMatch~\citep{barnes2009patchmatch}, were effective for small regions but struggled with complex textures and large gaps~\citep{jam2021comprehensive, liu2018image}. Deep learning advanced the field significantly, starting with Context Encoders~\citep{feature.learning.by.inpainting} and GAN-based methods like DeepFill~\citep{generative.inpainting, yu2019free}, improving content synthesis and coherence. Recent models such as LaMa~\citep{lama} use Fourier convolutional networks to address large masks. Diffusion models~\citep{NEURIPS2020_4c5bcfec}, notably Stable Diffusion~\citep{rombach2022high}, introduced iterative refinement capabilities, providing more flexible and structurally consistent inpainting compared to GANs~\citep{dhariwal2021diffusion}.


\vspace{3pt}
\noindent {\bf Diffusion models for image editing and inpainting.}
Beyond direct inpainting, diffusion models are widely used for image editing. SDEdit~\citep{meng2022sdedit} injects Gaussian noise and iteratively denoises, enabling semantic edits while preserving global structure. Noise inversion techniques~\cite{mokady2022null, miyake2024negativepromptinversionfastimage}, such as DDIM Inversion~\citep{song2020denoising}, further improve editing fidelity by enabling precise latent inference through deterministic reverse diffusion.
Inpainting-specific diffusion models like SDXL-Inpainting~\cite{podell2023sdxlimprovinglatentdiffusion} enhance image reconstruction by fine-tuning Stable Diffusion. Reference-based methods~\cite{tang2023realfill}, such as LeftRefill~\citep{cao2024leftrefill}, use diffusion models for reference-guided synthesis but struggle in regions distant from reference views.
Despite advancements, Stable Diffusion-based inpainting~\cite{inpaint3d} still suffers from inconsistent artifacts in scene-dependent contexts, causing multi-view inconsistencies problematic for 3D scenes~\citep{li2023diffusion}. This motivates our use of SDEdit and DDIM Inversion to preserve structural information and ensure multi-view coherence.

\vspace{3pt}
\noindent {\bf 3D scene inpainting.}
Existing 3D inpainting methods for NeRF~\cite{weder2022removing, spinnerf, shen2023nerfin, yin2023or, lin2024taming} typically adapt 2D models to NeRF’s implicit representation. For instance, SPIn-NeRF~\cite{spinnerf} employs perceptual loss to improve multi-view consistency. Reference-based methods~\cite{mirzaei2023reference, mirzaei2024reffusionreferenceadapteddiffusion, wang2024gscream} enhance consistency using reference images but remain limited to small-angle view rendering, restricting their use in 360° scenes. NeRFiller~\cite{weber2023nerfiller} iteratively refines consistency with grid prior but struggles with fine-grained textures due to image downsampling. InNeRF360~\cite{wang2023inpaintnerf360} handles 360° scenes via density hallucination but has limited scene utilization.
Gaussian Splatting-based methods like Gaussian Grouping~\cite{ye2023gaussian} inject semantic information, while InFusion~\cite{liu2024infusion} employs depth completion but requires manual view selection. GScream~\cite{scaffoldgs} integrates Scaffold-GS but faces difficulties in unbounded 360° scenes. Our method addresses these issues by enhancing multi-view consistency and depth-aware inpainting in 360° scenarios using Gaussian Splatting.

\begin{figure*}[t]
    \centering
    \includegraphics[width=1\linewidth]{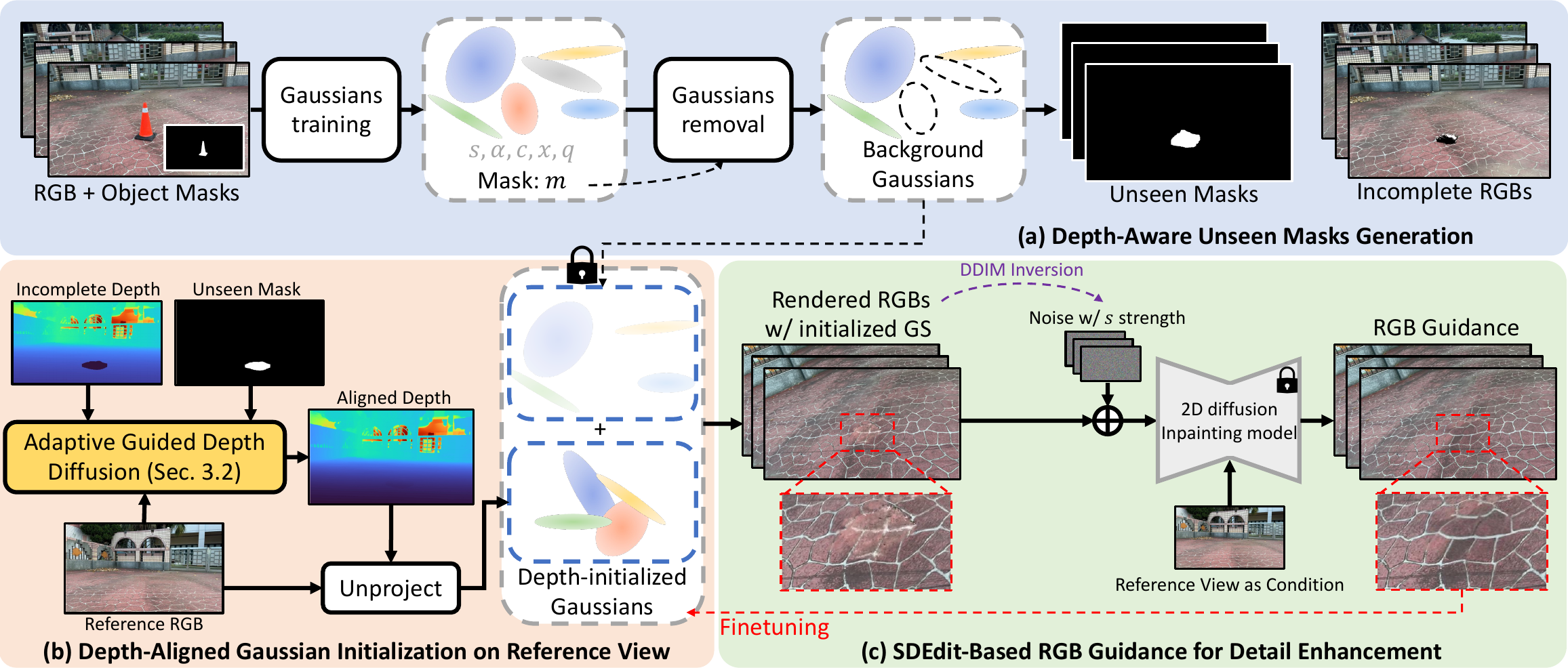}
    \vspace{-6mm}
    \caption{\textbf{Overview of our method.} Our approach takes multi-view RGB images and corresponding object masks as input and outputs a Gaussian representation with the masked objects removed. The pipeline consists of three main stages: (a) Depth-Aware Unseen Masks Generation to identify truly occluded areas, referred to as the ``unseen region'', (b) Depth-Aligned Gaussian Initialization on Reference View to fill unseen regions with initialized Gaussian containing reference RGB information after object removal, and (c) SDEdit-Based RGB Guidance for Detail Enhancement, which enhances fine details using an inpainting model while preserving reference view information. Instead of applying SDEdit with random noise, we use DDIM Inversion on the rendered initial Gaussians to generate noise that retains the structure of the reference view, ensuring multi-view consistency across all RGB Guidance.}
    \label{fig:pipeline}
    \vspace{-1mm}
\end{figure*}

\section{Method}
\label{sec:method}


Our method processes multi-view RGB images $\left\{ I_n \right\}$ and object masks $\left\{ M_n \right\}$, $n \in \left[1..N \right]$, to produce an inpainted Gaussian representation with removed objects. Occluded regions (unseen regions~\cite{ye2023gaussian}) are consistently inpainted across views. As shown in~\cref{fig:pipeline}, the process includes training a masked Gaussian using object masks, removing objects, and applying (a) Depth-Aware Unseen Mask Generation (\cref{sec:unseen}), (b) Reference View Initial Gaussians Alignment
(\cref{sec:depth_initial}), and (c) SDEdit for Detail Enhancement (\cref{sec:sdedit}). This pipeline ensures consistent texture propagation in unbounded scenes, achieving high-quality 3D inpainting.

\subsection{Depth-Aware Unseen Mask Generation} \label{sec:unseen}

Accurate identification of inpainting regions is critical for scene consistency and optimal use of background information. To generate the unseen mask for a view, it is necessary to differentiate between (1) the background visible across multiple views and (2) the unseen region occluded in all views, requiring inpainting.



\begin{figure}[t]
    \centering
    \includegraphics[width=1\linewidth]{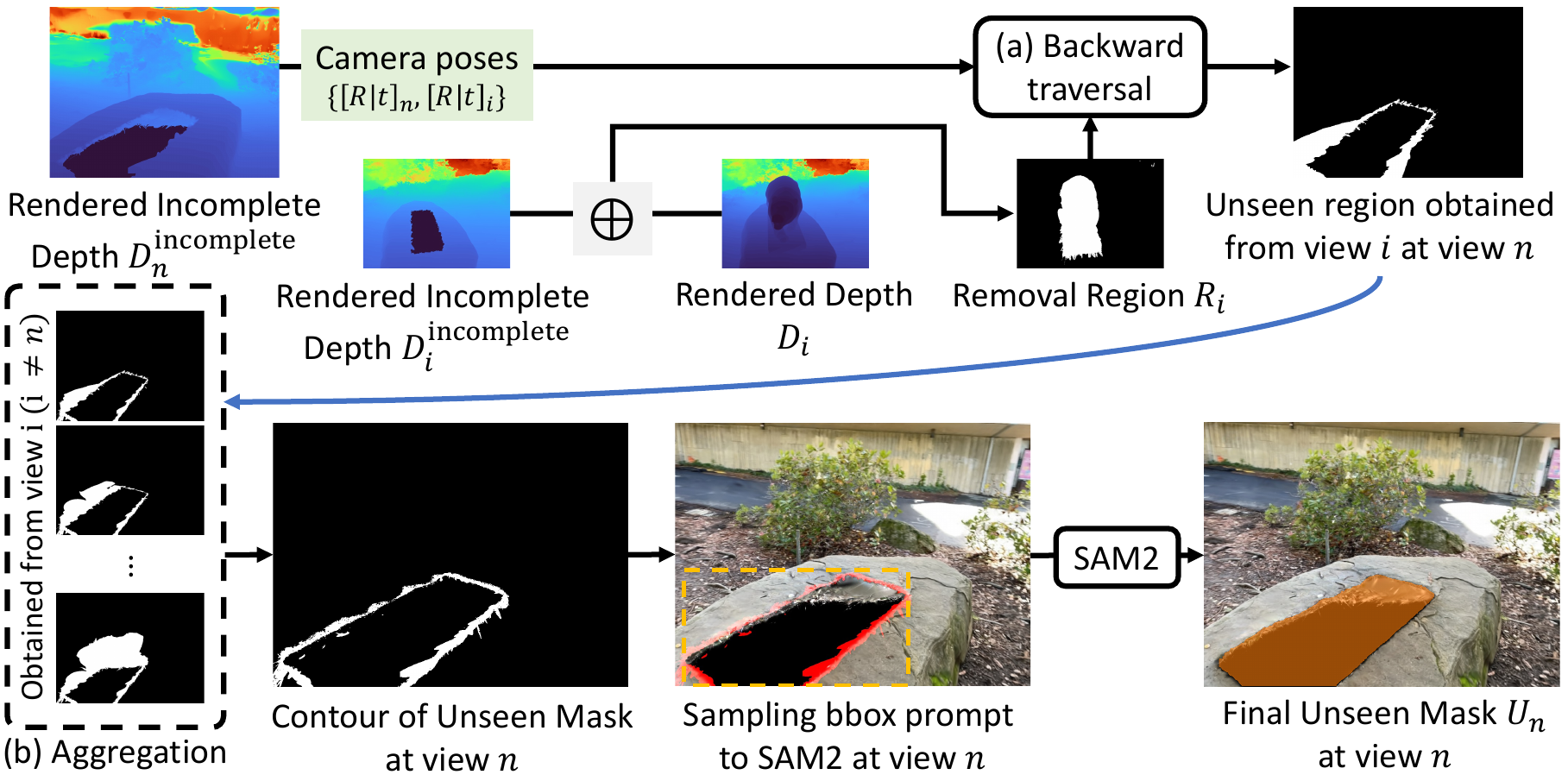}
    \vspace{-6mm}
    \caption{\textbf{Overview of the Unseen Mask Generation Process using Depth Warping.} To obtain the unseen mask for view $n$, we calculate the pixel correspondences between the view $n$ and all other views $i$ by using the rendered incomplete depth $D_{n}^{\text{incomplete}}$. For each view $i$, the removal region $R_i$ is backward traversal to view $n$ to align occlusions. We then aggregate the results from multiple views, averaging and applying a threshold to produce the initial contour of the unseen mask. This contour is subsequently converted into a bounding box prompt for SAM2~\cite{ravi2024sam2}, which refines the unseen mask to its final version for view $n$.}
    \label{fig:unseen_masks}
    \vspace{-1mm}
\end{figure}

A naive approach to detecting unseen masks with SAM2~\cite{ravi2024sam2} involves manually selecting the first view and propagating prompts across other views. However, SAM2 struggles to consistently detect unseen regions without refinement, often revealing parts of the background or inside objects. To address this, our method employs depth warping to generate bounding box prompts for each view (\cref{fig:unseen_masks}), ensuring accurate, fully automated unseen region detection.



\vspace{3pt}
\noindent {\bf Depth warping for generating bbox prompt to SAM2.}
To refine the unseen mask, we employ a depth-warping technique, as illustrated in \cref{fig:unseen_masks}. For each view $n$, we compute:
\begin{small}
\begin{equation}
R_{i \rightarrow n} = \mathcal{W}_{\text{traverse}}(R_{i}, D_n^{\text{incomplete}}, T_{n \rightarrow i}),
\end{equation}
\end{small}
where $\mathcal{W}_{\text{traverse}}$ includes forward warping from view $n$ to $i$ and backward traversal to map the removal region back to $n$. $R_i$ is the removal region mask for view $i$, derived from depth differences. $D_{n}^{\text{incomplete}}$ is the incomplete depth map for view $n$, and $T_{n \rightarrow i}$ is the transformation from view $n$ to $i$.  

The unseen mask contour for view $n$ is obtained by aggregating warped removal regions and applying thresholding:
\begin{small}
\begin{equation}
C_n = \theta \left( \frac{1}{K} \sum_{i=1}^K R_{i \rightarrow n} \right) \cap R_{n},
\end{equation}
\end{small}
where $C_n$ is the contour of the unseen mask, $K$ is the number of views, and $\theta$ is a thresholding function. A bounding box $\text{bbox}(C_n)$ is created as a prompt for SAM2~\cite{ravi2024sam2} to generate the final unseen mask:
\begin{small}
\begin{equation}
U_n = \text{SAM2}(\text{bbox}(C_n)).
\end{equation}
\end{small}
This mask $U_n$ guides the inpainting process, focusing on areas needing reconstruction while preserving original scene information.

\subsection{Reference View Initial Gaussians Alignment} \label{sec:depth_initial}
After performing object removal and generating the unseen mask, similar to CorrFill~\cite{liu2025corrfill}, we select a reference view called \( V_{\text{ref}} \), which can render an incomplete RGB image and depth. We then apply RGB inpainting to the incomplete RGB image of \( V_{\text{ref}} \) and denote it as $I_{\text{ref}}$. 
To maximize cross-view consistency, we project the reference RGB image into 3D space using depth estimates of $I_{\text{ref}}$, which is obtained through Adaptive Guided Depth Diffusion. This 3D projection serves two critical purposes: It guides the SDEdit-based RGB detail enhancement and initializes point positions for Gaussian fine-tuning. Accurate depth alignment is, therefore, fundamental to our pipeline, as it directly determines the precision of these initial point positions.

\vspace{3pt}
\noindent {\bf Adaptive Guided Depth Diffusion (AGDD).}
Aligning \textit{estimated depth} with \textit{existing depth} is challenging due to monocular depth estimation~\cite{ke2023repurposing}'s scale ambiguity and non-metric representation across coordinate systems. This challenge intensifies in 360° unbounded scenes, where large viewpoint changes hinder alignment. Traditional scale-shift optimization often yields suboptimal results, while depth-completion models demand costly fine-tuning. Our AGDD refines GDD~\cite{yu2024wonderworld} by addressing over-alignment issues, particularly where depth transitions from small to large values, which exaggerates disparities in distant regions and inflates loss values. To mitigate this, we introduce an adaptive loss $L_{\text{adaptive}}$ that balances alignment, preventing distant regions from dominating and yielding more accurate depth estimates.

\begin{figure}[t]
    \centering
    \includegraphics[width=1\linewidth]{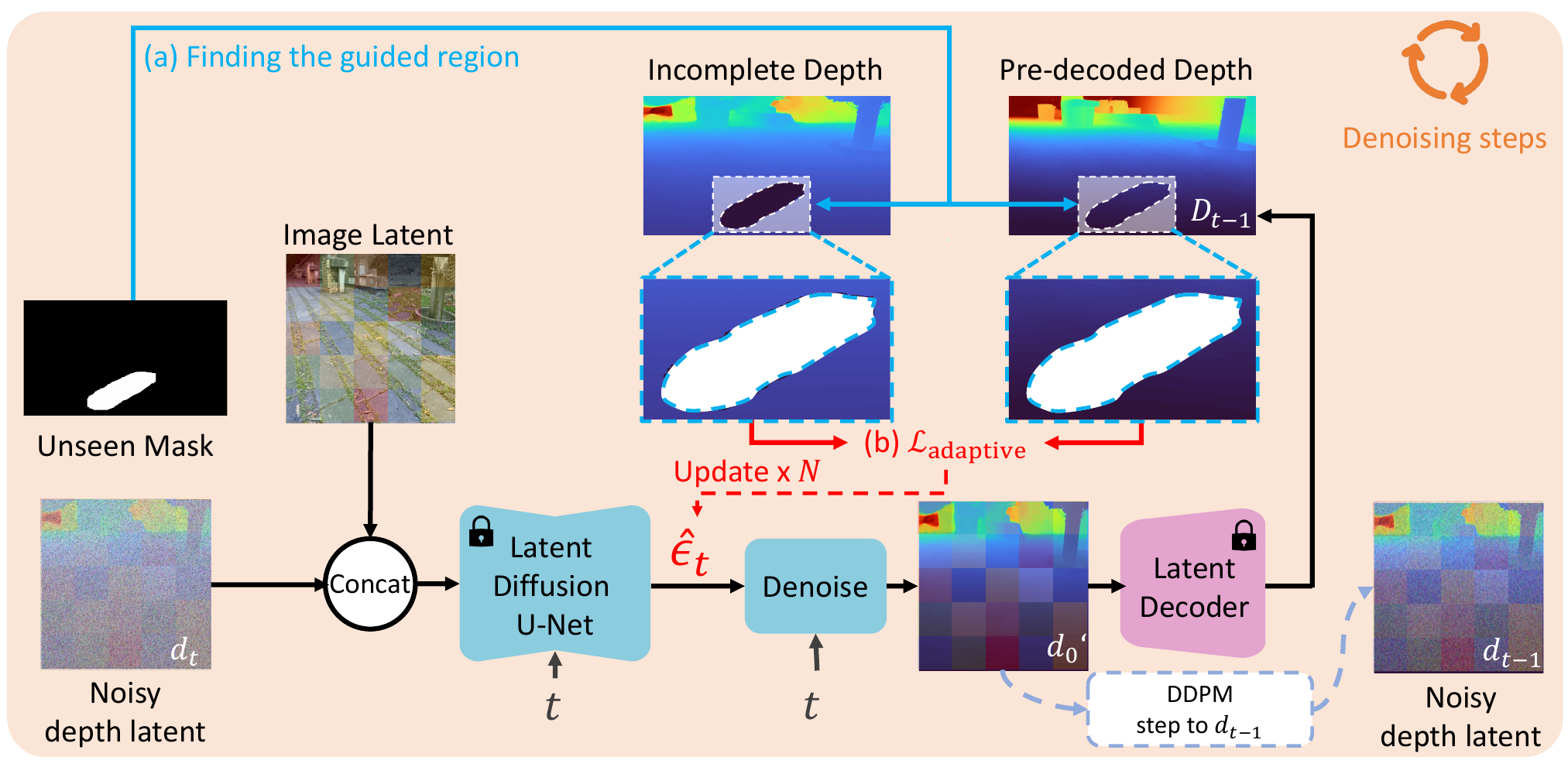}
    \vspace{-6mm}
    \caption{
    \textbf{Overview of Adaptive Guided Depth Diffusion (AGDD).}
    The framework takes image latent, incomplete depth, and unseen mask as inputs to generate aligned depth estimates. (a) The guided region is identified by dilating the unseen mask and subtracting the original mask. (b) At each timestep $t$, adaptive loss $\mathcal{L}_\text{adaptive}$ is computed between the pre-decoded and incomplete depth to update the noise input $\hat{\epsilon}_t$. This repeats $N$ times before advancing to the next denoising step, ensuring the estimated depth aligns with the incomplete depth distribution in the guided region.
    }
    \label{fig:AGDD}
    \vspace{-1mm}
\end{figure}

The framework is shown in~\cref{fig:AGDD}. Following the standard denoising process of Marigold~\cite{ke2023repurposing}, we initialize with a latent representation perturbed by full-strength Gaussian noise, denoted as $d_t$, and generate aligned depth $D_{\text{aligned}}$ = $\text{Decoder}(d_0)$ using a VAE decoder, where the latent $d_{0}$ is obtained by recursive denoising step $d_{t-1} = \text{Denoise}(d_t, t, \hat{\epsilon}_t)$. The $\hat{\epsilon}_t$ is derived by updating the original noise through the calculation of adaptive loss $L_{\text{adaptive}}$ between the pre-decoded estimated depth $D_{t-1}$ and the existing incomplete depth $D_{\text{incomplete}}$. Note that $D_{t-1}$ is obtained by decoding $d_{0}^{'}$, which is the model's estimation of the fully denoised latent at timestep $0$ when predicted from the noisy state at timestep $t-1$. This adaptive loss refines $\hat{\epsilon}_t$ to ensure that the estimated depth aligns with the existing incomplete depth during denoising. The optimization process is described as follows:
\begin{small}
\begin{equation} 
d_{t-1} = \text{Denoise}(d_t, t, \hat{\epsilon}_t) 
\end{equation} 
\end{small}
\begin{small}
\begin{equation} 
\hat{\epsilon}_t = \text{UNet}(d_t, I_{\text{scene}}, t) - \alpha \cdot \nabla \mathcal{L}_{\text{adpative}}
\end{equation} 
\end{small}
where \( \alpha \) is the learning rate for the optimization. We define a bounding box $\mathcal{B}$ around the unseen region and introduce a threshold $\delta$ to downweight errors for distant points. The adaptive loss $\mathcal{L}_{\text{adaptive}}$ between the pre-decoded estimated depth $D_{t-1}$ and the incomplete depth $D_{\text{incomplete}}$ is computed as follows:
\begin{small}
\begin{equation}
M_{\text{guide}}(x, y) = 
\begin{cases} 
1 & \text{if } (x, y) \in \mathcal{B} \setminus U \\
0 & \text{otherwise},
\end{cases}
\end{equation}
\end{small}
\begin{small}
\begin{equation}
\mathcal{L}_{\text{adaptive}} = \sum_{(x,y)} M_{\text{guide}}(x,y) \cdot \mathcal{L}(D_{t-1}, D_{\text{incomplete}})(x, y),
\end{equation}
\end{small}
\begin{small}
\begin{equation}
\mathcal{L}(d_1, d_2) = 
\begin{cases} 
\frac{1}{2} (d_1 - d_2)^2 & \text{if } |d_1 - d_2| < \delta \\
\delta \cdot |d_1 - d_2| - \frac{1}{2} \delta^2 & \text{otherwise,}
\end{cases}
\end{equation}
\end{small}
where $M_{\text{guide}}(x, y)$ is a mask function indicating if a pixel $(x, y)$ is within the bounding box $\mathcal{B}$ but not in the unseen mask U. At each denoising step, we update the noise over $N$ iterations. Instead of directly optimizing the noise using L2 loss~\cite{yu2024wonderworld}, this loss ensures that the updated noise input to the denoiser enables it to generate an estimated depth that aligns with the incomplete guided depth. This enables the AGDD output to achieve accurate alignment in regions adjacent to unseen areas, which is more appropriate for depth inpainting scenarios while also operating in a zero-shot manner.

\vspace{2pt}
\noindent {\bf Initializing Gaussians in unseen regions.}
With the aligned depth $D_\text{aligned}^{\text{ref}}$ of the reference view, we proceed to initialize new Gaussians in the unseen regions. First, we unproject the inpainted RGB of the reference view with $D_\text{aligned}^\text{ref}$ to 3D space, focusing on the unseen regions identified by the unseen mask. This unprojection takes into account the camera's intrinsic parameters. For each pixel $(u, v)$ in the unseen region where $U_\text{final}(u, v) = 1$, we compute the 3D point $P = (X, Y, Z)$ as $Z = D_\text{aligned}^\text{ref}(u, v)$, $X = (u - c_x) \cdot Z / f_x$, $Y = (v - c_y) \cdot Z / f_y,$,
where $(f_x, f_y)$ are the focal lengths in pixels and $(c_x, c_y)$ are the principal point offsets. This process gives us a set of initial 3D points $P$. These points are then used to initialize new Gaussians in the unseen regions, inheriting color from the reference view. Existing background Gaussians, unaffected by object removal, remain fixed during initialization and optimization. These initialized Gaussians are crucial for the subsequent process of generating guided inpaint RGB images and optimization.

\subsection{SDEdit for Detail Enhancement}  
\label{sec:sdedit}  

After initializing Gaussians in unseen regions, we aim to obtain the inpainted RGB guidance with fine details while ensuring multi-view consistency, which further refines our initial Gaussians during fine-tuning. Inspired by SDEdit~\citep{meng2022sdedit}, we refine the rendered initial Gaussians by adding scaled noise proportional to a strength factor \( s \), ensuring that the inpainting model retains structural information from the reference view while allowing for detail refinement across multiple perspectives. We further find that instead of injecting random Gaussian noise, applying DDIM Inversion~\cite{song2020denoising} to the rendered initial Gaussians better preserves their structural information during the denoising process. This approach allows the diffusion inpainting model to reconstruct missing details while maintaining alignment with the reference view, ensuring that inpainted regions integrate seamlessly into the scene (see~\cref{fig:ablation_depth_alignment2}).

Specifically, given a rendered training view \( I_{\text{init}} \), we first obtain its corresponding noise representation via DDIM Inversion, capturing the essential structure of the reference view in the latent space. Instead of inverting fully to \( t_0 \), we compute an intermediate timestep \( t_{\text{inv}} \) based on the noise strength \( s \):
\begin{small}
\begin{equation}  
t_{\text{inv}} = T (1 - s),  
\end{equation}  
\end{small}
where \( T \) is the total number of timesteps in the diffusion process, and \( s \) controls the noise strength. We then perform DDIM Inversion to obtain the noise representation at \( t_{\text{inv}} \):
\begin{small}
\begin{equation}  
\epsilon_{\text{inv}} = \text{DDIM-Invert}(I_{\text{init}}, t_{\text{inv}}).  
\end{equation}  
\end{small}

Next, we denoise this noise using a 2D diffusion inpainting model, conditioned on the reference view \( I_{\text{ref}} \), ensuring that the reconstructed details align with the global scene while maintaining consistency across views:
\begin{small}
\begin{equation}  
I_{\text{guided}} = \text{Denoise}(\epsilon_{\text{inv}}, \text{condition} = I_{\text{ref}}, t_{\text{inv} \rightarrow }0).  
\end{equation}  
\end{small}
By inverting to a noise level corresponding to strength \( s \), this step ensures that the inpainting model refines details while maintaining geometric consistency with the reference view. Unlike traditional SDEdit, which applies random noise addition before denoising, our approach leverages DDIM Inversion to obtain structured noise that aligns with the scene, preventing hallucinated details that could disrupt multi-view coherence.  

The resulting guided inpainted RGBs are then used as supervision for Gaussian fine-tuning, updating only the unprojected Gaussians from Sec.~\ref{sec:depth_initial}. The final reconstruction is optimized using a combination of L1, SSIM, and LPIPS~\citep{zhang2018unreasonable} losses:  
\begin{small}
\begin{equation}  
\mathcal{L} = (1 - \lambda_\text{SSIM}) \mathcal{L}_1 + \lambda_\text{SSIM} \mathcal{L}_\text{SSIM} + \lambda_\text{LPIPS} \mathcal{L}_\text{LPIPS}.  
\end{equation}  
\end{small}

\subsection{Implementation Details}

We use the 2D Gaussian Splatting~\cite{huang20242d} codebase for Gaussian representation to obtain accurate rendered depth, with SAM2 generating object masks on the first frame for each training view. Masked Gaussians enable effective object removal due to their explicit representation. We set the aggregation threshold of \(\theta\) to 0.6 in unseen mask generation. In AGDD, incomplete depth are normalized to match Marigold's~\cite {ke2023repurposing} depth. With $N$ set to 8, the denoised result is then unnormalized back to its original scale. The entire inference process takes approximately 1 minute on an RTX 4090 GPU. The noise strength of SDEdit \(s = 0.85\) balances initial point retention, as shown in our ablation study. We condition the generation on the reference view using LeftRefill~\cite{cao2024leftrefill}. During Gaussian fine-tuning, we run 10,000 iterations with \(\lambda_{\text{SSIM}} = 0.8\) and \(\mathcal{L}_{\text{LPIPS}} = 0.5\).

 


\begin{figure*}[t]
    \centering
    \includegraphics[width=1\linewidth]{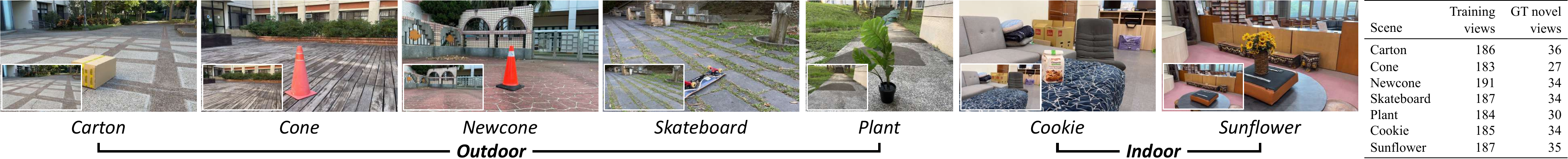}
    \vspace{-6mm}
    \caption{\textbf{Overview of the 360-USID dataset.} Sample images from each scene, including five outdoor scenes (Carton, Cone, Newcone, Skateboard, Plant) and two indoor scenes (Cookie, Sunflower). (\emph{Bottom right}) The table shows statistics for each scene, including the number of training views and ground truth (GT) novel views. The dataset provides a diverse range of environments for evaluating 3D inpainting methods in both indoor and outdoor settings.}
    \label{fig:our_dataset}
    \vspace{-3mm}
\end{figure*}

\begin{figure}[t]
    \centering
    \includegraphics[width=1\columnwidth]{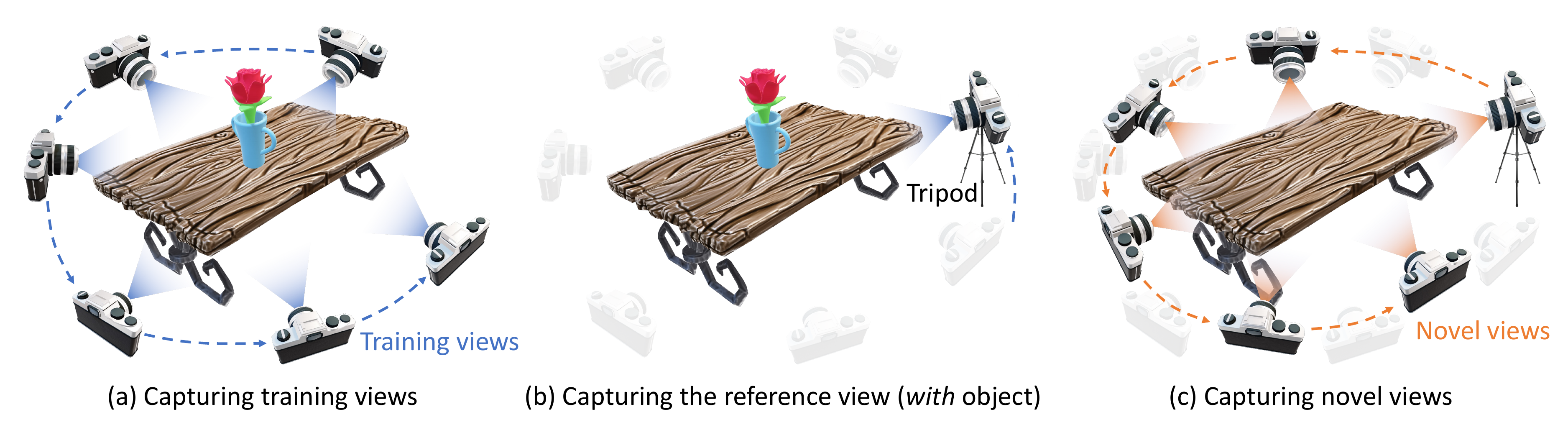}
    \vspace{-6mm}
    \caption{\textbf{Illustration of the data capture process for the 360-USID dataset.} (a) Capturing training views: Multiple images are taken around the object in the scene. (b) Capturing the reference view: A camera is mounted on a tripod to capture a fixed reference view (with an object). (c) Capturing novel views: After removing the object, additional images are taken from various viewpoints, including one from the same tripod position as the reference image.}
    \label{fig:dataset_capture}
\end{figure}

\section{360$^{\circ}$ Unbounded Scenes Inpainting Dataset}
\label{sec:dataset}
To address the lack of reference-based 360° inpainting datasets, we introduce the 360° Unbounded Scenes Inpainting Dataset (360-USID), consisting of seven scenes with training views (RGB images and object masks), novel testing views (inpainting ground truth), and a reference view (without objects) for evaluating with other reference-based methods.

\vspace{3pt}
\noindent {\bf Dataset collection protocol.}
We developed a protocol using a standard camera to create this dataset, as simultaneously capturing multi-view photos with and without objects typically requires specialized equipment. Our protocol, illustrated in~\cref{fig:dataset_capture}, consists of:
\begin{enumerate}
\item Positioning an object (\emph{e.g.} a vase) on a textured surface within a 360° unbounded scene. Training views are captured in two complete circular trajectories around the object - the first focuses primarily on the object, while the second maximizes background coverage to ensure comprehensive scene capture.
\item Securing the camera on a tripod and capturing a reference view from a fixed position and orientation.
\item After object removal, capturing novel views from both the fixed tripod position and additional positions distinct from training trajectories for ground truth evaluation.
\end{enumerate}
To ensure high-quality captures, we record video at 4K 60fps with stabilized camera settings and extract the sharpest frames using the variance of the Laplacian method. Each scene comprises 180$\sim$200 training views and approximately 30 testing views for quantitative evaluations. Consistent lighting is maintained throughout to minimize shadow variations between reference and testing images

\vspace{3pt}
\noindent {\bf Data preprocessing and pose estimation.}
\label{sec:data_preprocessing}
Our processing pipeline begins with using COLMAP~\citep{schoenberger2016sfm,schoenberger2016mvs} or similar SfM pipelines like hloc~\citep{sarlin2019coarse,sarlin2020superglue} to compute a shared 3D coordinate space for both training and novel views. We then generate object masks for training views using SAM2~\citep{ravi2024sam2} and mask out object regions in COLMAP reconstruction. After obtaining camera poses, we process the training images with NeRF/3DGS inpainting methods and render novel views for comparison against ground truth. Finally, we refine testing views by training a masked-3DGS model and selecting optimal frames based on PSNR scores computed outside object regions, yielding approximately 30 high-quality test views per scene. The resulting dataset provides a comprehensive benchmark for evaluating 360° inpainting methods across diverse scenes and viewpoints, with particular attention to view consistency and geometric accuracy.

\vspace{3pt}
\noindent {\bf Scene descriptions.}
Our 360-USID dataset, shown in~\cref{fig:our_dataset}, contains seven diverse scenes: five outdoor (Carton, Cone, Newcone, Plant, Skateboard) and two indoor (Cookie, Sunflower). Each scene includes 180-200 training images at 3840$\times$2160 resolution (Plant at 1920$\times$1440), 30 ground truth testing images, and one reference image without objects. Scenes are downscaled to 960$\times$540 for evaluation, providing a comprehensive benchmark for testing 3D inpainting methods across varied real-world environments.

\section{Experiments}

\subsection{Experimental setup}

\noindent {\bf Datasets.}
We evaluate on two 360° unbounded scene datasets:
(1) \textbf{360-USID (Ours)}: A new dataset of 7 scenes (3 indoor, 4 outdoor) for evaluating 360° inpainting, with 200-300 training views containing objects, around 30 test views without objects, and 1 reference. All images are processed at 960px width to preserve details for quantitative evaluation.
(2) \textbf{Other-360~\citep{barron2022mip}} We collect additional 6 standard 360° unbounded scene datasets from NeRF\cite{mildenhall2020nerf}, MipNeRF-360\citep{barron2022mip} and Instruct-NeRF2NeRF\cite{haque2023instruct} for qualitative evaluation at 1/4 resolution, with frame 0 as reference for all methods.




\vspace{3pt}
\noindent {\bf Metrics.}
We evaluate our method using two complementary metrics: LPIPS (Learned Perceptual Image Patch Similarity)~\citep{zhang2018unreasonable} for perceptual quality and PSNR (Peak Signal-to-Noise Ratio) for reconstruction accuracy. Following SPIn-NeRF~\citep{spinnerf}, we compute these metrics only within object masks to focus on inpainting quality. While both metrics are used for 360-USID, which has ground truth, only qualitative assessment is possible for Other-360. Additional evaluation results are provided in supplementary materials.

\begin{table*}[t]
\centering
\caption{\textbf{Quantitative comparison of 360° inpainting methods on the 360-USID dataset.} \textcolor{red}{Red} text indicates the best, and \textcolor{blue}{blue} text indicates the second-best performing method.}
\label{tab:quantitative}
\vspace{-3mm}
\resizebox{\textwidth}{!}{%
\begin{tabular}{l|ccccccc|c}
\toprule
PSNR $\uparrow$ / LPIPS $\downarrow$ & Carton & Cone & Cookie & Newcone & Plant & Skateboard & Sunflower & Average \\
\midrule
SPIn-NeRF~\cite{spinnerf} & 16.659 / 0.539 & 15.438 / 0.389 & 11.879 / 0.521 & \textcolor{blue}{17.131} / 0.519 & 16.850 / 0.401 & 15.645 / 0.675 & 23.538 / 0.206 & \textcolor{blue}{16.734} / 0.464 \\
2DGS~\cite{huang20242d} + LaMa~\cite{lama} & 16.433 / 0.499 & 15.591 / \textcolor{blue}{0.351} & 11.711 / 0.538 & 16.598 / 0.670 & 14.491 / 0.564 & 15.520 / 0.639 & 23.024 / 0.194 & 16.195 / 0.494 \\
2DGS~\cite{huang20242d} + LeftRefill~\cite{cao2024leftrefill} & 15.157 / 0.567 & \textcolor{red}{16.143} / 0.372 & \textcolor{blue}{12.458} / 0.526 & 16.717 / 0.677 & 12.856 / 0.666 & \textcolor{blue}{16.429} / 0.634 & 24.216 / 0.181 & 16.282 / 0.518 \\
LeftRefill~\cite{cao2024leftrefill} & 14.667 / 0.560 & 14.933 / 0.380 & 11.148 / 0.519 & 16.264 / \textcolor{blue}{0.448} & 16.183 / 0.463 & 14.912 / \textcolor{blue}{0.572} & 18.851 / 0.331 & 15.280 / 0.468 \\
Gaussian Grouping~\cite{ye2023gaussian} & \textcolor{blue}{16.695} / \textcolor{blue}{0.502} & 14.549 / 0.366 & 11.564 / 0.731 & 16.745 / 0.533 & 16.175 / 0.440 & 16.002 / 0.577 & 20.787 / 0.209 & 16.074 / 0.480 \\
GScream~\cite{wang2024gscream} & 14.609 / 0.587 & 14.655 / 0.476 & 12.733 / \textcolor{red}{0.429} & 13.662 / 0.605 & 16.238 / 0.437 & 12.941 / 0.626 & 18.470 / 0.436 & 14.758 / 0.514 \\

Infusion~\cite{liu2024infusion} & 14.191 / 0.555 & 14.163 / 0.439 & 12.051 / 0.486 & 9.562 / 0.624 & 16.127 / 0.406 & 13.624 / 0.638 & 21.195 / 0.238 & 14.416 / 0.484 \\
AuraFusion360 (Ours) w/o SDEdit & 13.731 / 0.477 & 14.260 / 0.390 & 12.332 / 0.445 & 16.646 / 0.460 & \textcolor{blue}{17.609} / \textcolor{red}{0.319} & 15.107 / 0.580 & \textcolor{blue}{24.884} / \textcolor{red}{0.170} & 16.367 / \textcolor{blue}{0.406} \\
AuraFusion360 (Ours) & \textcolor{red}{17.675} / \textcolor{red}{0.473} & \textcolor{blue}{15.626} / \textcolor{red}{0.332} & \textcolor{red}{12.841} / \textcolor{blue}{0.434} & \textcolor{red}{17.536} / \textcolor{red}{0.426} & \textcolor{red}{18.001} / \textcolor{blue}{0.322} & \textcolor{red}{17.007} / \textcolor{red}{0.559} & \textcolor{red}{24.943} / \textcolor{blue}{0.173} & \textcolor{red}{17.661} / \textcolor{red}{0.388} \\

\bottomrule
\end{tabular}%
}
\end{table*}



\begin{figure*}[t]
    \centering
    \includegraphics[width=1\linewidth]{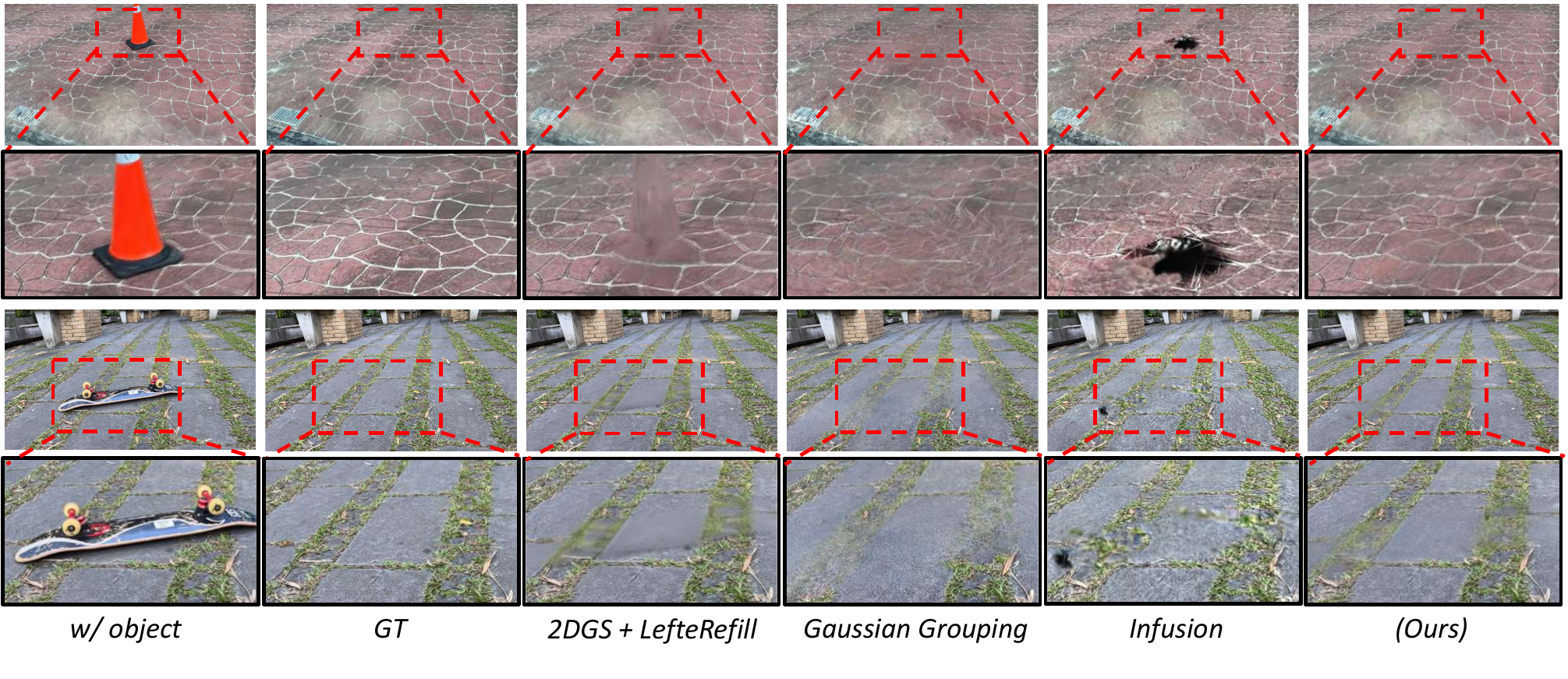}
    \vspace{-9mm}
    \caption{\textbf{Visual Comparison on our 360-USID dataset.} We compare our method against state-of-the-art approaches including Gaussian Grouping~\citep{ye2023gaussian}, 2DGS + LeftRefill, and Infusion~\citep{liu2024infusion}. While Gaussian Grouping struggles with misidentifying unseen regions, leading to floating artifacts, and 2DGS + LeftRefill faces view consistency issues, our method successfully maintains geometric consistency and preserves fine details across different viewpoints. Ground truth (GT) is shown for reference, and the original scene with an object is provided in the first row for comparison.}
    \label{fig:visual}
\end{figure*}

\subsection{Comparisons with State-of-the-Art Methods}

\noindent {\bf Quantitative comparisons.}
We evaluate AuraFusion360 against state-of-the-art approaches on the 360-USID dataset. \cref{tab:quantitative} shows PSNR and LPIPS scores across different scenes. Our method consistently outperforms existing approaches. SPIn-NeRF~\cite{spinnerf}\footnotemark and Infusion~\cite{liu2024infusion} struggle with 360° consistency, while Gaussian Grouping~\citep{ye2023gaussian} misidentifies the unseen region, causing significant floating artifacts. GScream~\cite{wang2024gscream} fails to properly remove objects, and LeftRefill~\cite{cao2024leftrefill} improves but still falls short in 360° environments. 2DGS + LaMa~\citep{lama} and 2DGS + LeftRefill outperform 2D methods but face view consistency challenges. Our method achieves the highest PSNR score and the lowest average LPIPS, indicating superior perceptual quality and better similarity to the ground truth. The performance gap is especially noticeable in scenes with complex geometry or large removed objects, demonstrating our method's ability to leverage multi-view information and maintain 360° consistency. The code for InNeRF360~\cite{wang2023inpaintnerf360} could not be successfully executed, and~\cite{mirzaei2023reference} did not provide code, so we were unable to compare our method with theirs.
\footnotetext{We implement SPin-NeRF's method on the 2D Gaussian Splatting codebase to extend its capabilities to 360° unbounded scenes.}


\vspace{3pt}
\noindent {\bf Qualitative visual comparisons.}
\cref{fig:visual} compares our AuraFusion360 method against state-of-the-art approaches on challenging scenes from the 360-USID dataset. Our method excels in maintaining view consistency and preserving fine details in 360° unbounded environments. Additional qualitative results on other 360 datasets and failure cases are provided in the supplementary material. 

\begin{table}[t]
\centering
\small
\caption{\textbf{Ablation study of our AuraFusion360.}
} 
\label{tab:ablation}
\vspace{-3mm}
\begin{tabular}{cc|cc}
\toprule
Depth init. & SDEdit strength & PSNR $\uparrow$ & LPIPS $\downarrow$ \\
 (\cref{sec:depth_initial}) & (\cref{sec:sdedit}) &  &  \\
\midrule
  & 0.85 &  16.638 & 0.456\\
\checkmark  & 0.5 & 17.646 & 0.393 \\
\checkmark  & 1.0 & 17.512 & 0.391 \\
\checkmark  & 0.85 & \textbf{17.661} & \textbf{0.388} \\
\bottomrule
\end{tabular}%
\end{table}




\subsection{Ablation Studies}
To evaluate the effectiveness of each component in our AuraFusion360 method, we conduct a series of ablation studies. \cref{tab:ablation} presents the quantitative results of these studies. 

\begin{figure}[t]
    \centering
    \includegraphics[width=1\linewidth]{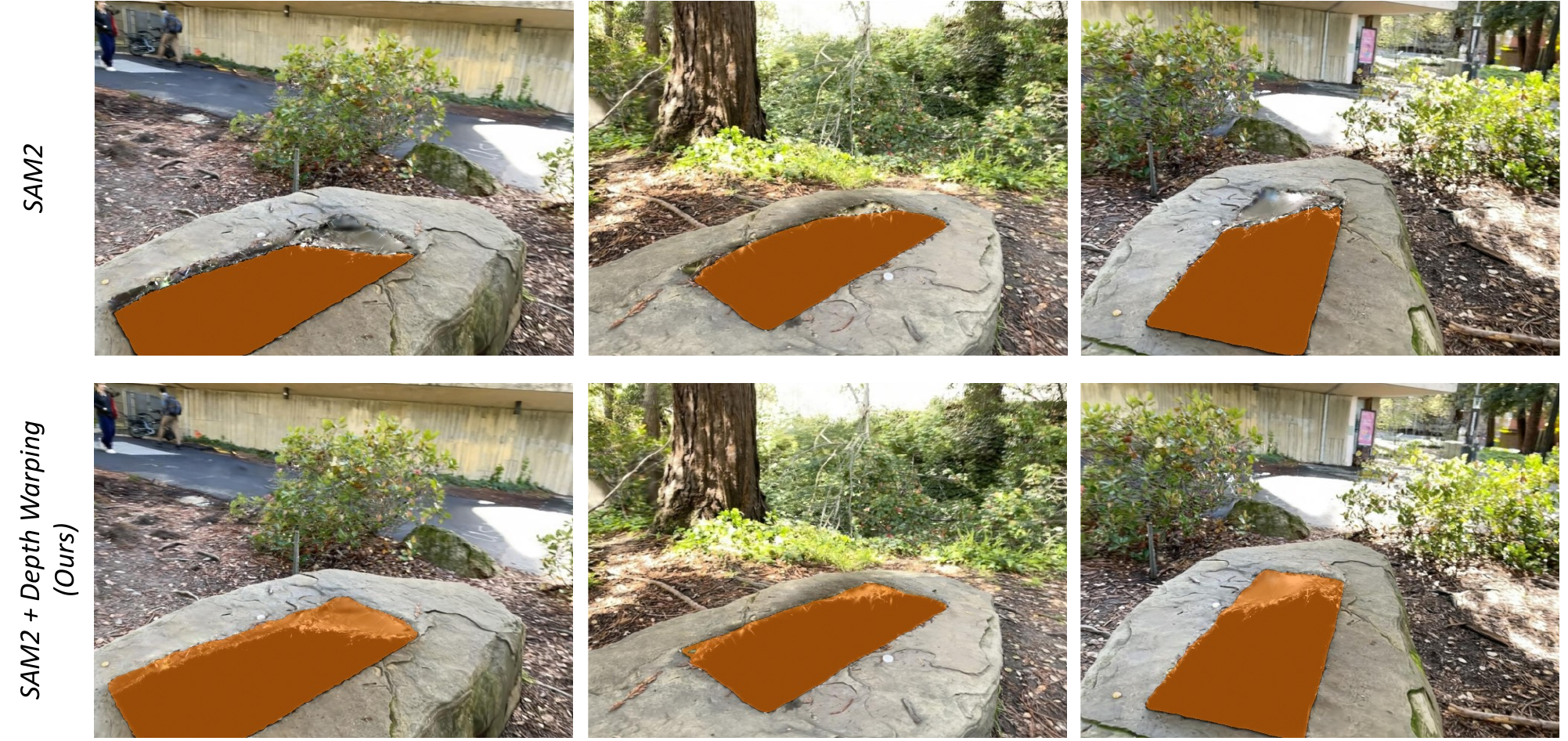}
    \vspace{-6mm}
    \caption{\textbf{Visual comparison of unseen mask generation method.} Our method enables SAM2~\cite{ravi2024sam2} to generate more accurate predictions for each view without the need for manually provided prompts, as the bounding box prompts are automatically generated through depth warping.} 
    \label{fig:ablation_unseen}
\end{figure}
\begin{figure}[t]
    \centering
    \includegraphics[width=1\linewidth]{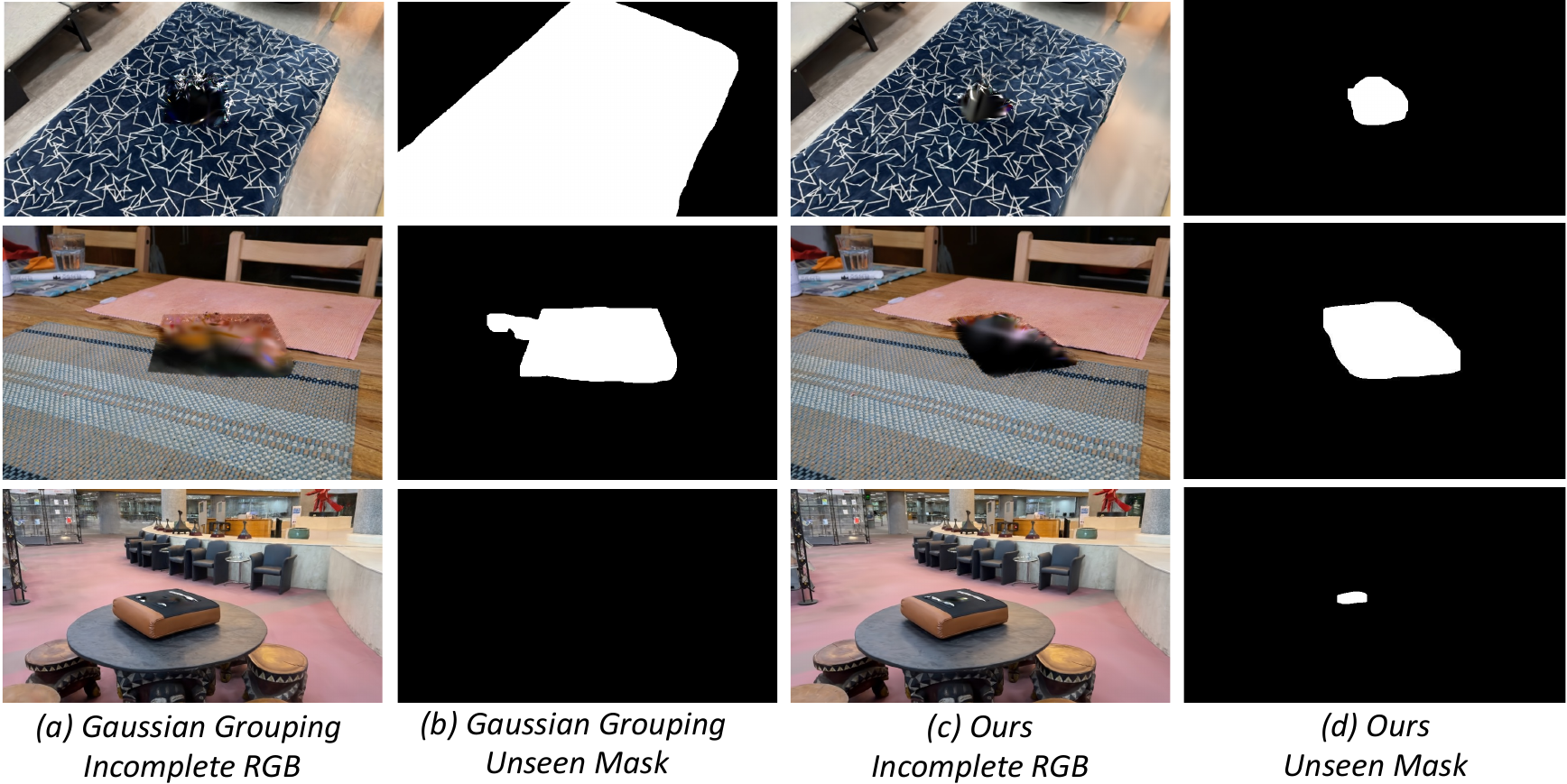}
    \vspace{-6mm}
    \caption{
    \textbf{Compared Unseen Mask w/ Gaussian Grouping.} 
    Gaussian Grouping~\cite{ye2023gaussian} uses a video tracker~\cite{cheng2023tracking} and the ``black blurry hole'' prompt for DEVA~\cite{cheng2023tracking} to track the unseen region. However, this can result in tracking errors, affecting inpainting. In contrast, our geometry-based approach uses depth warping to estimate the unseen region's contour, reducing segmentation errors.
    } 
    \label{fig:visual_unseen}
\end{figure}

\vspace{3pt}
\noindent {\bf Unseen mask generation.}
We compared our unseen mask generation method with SAM2~\cite{ravi2024sam2} and Gaussian Grouping~\cite{ye2023gaussian} tracker in ~\cref{fig:ablation_unseen} and ~\cref{fig:visual_unseen}. Our approach significantly improves inpainting quality, particularly in areas occluded from multiple views. The unseen masks identify truly occluded regions, leading to more accurate and consistent inpainting results. This is especially noticeable in scenes with complex geometries, where object masks alone may not capture all necessary information for effective inpainting.

\begin{figure}[t]
    \centering
    \includegraphics[width=1\linewidth]{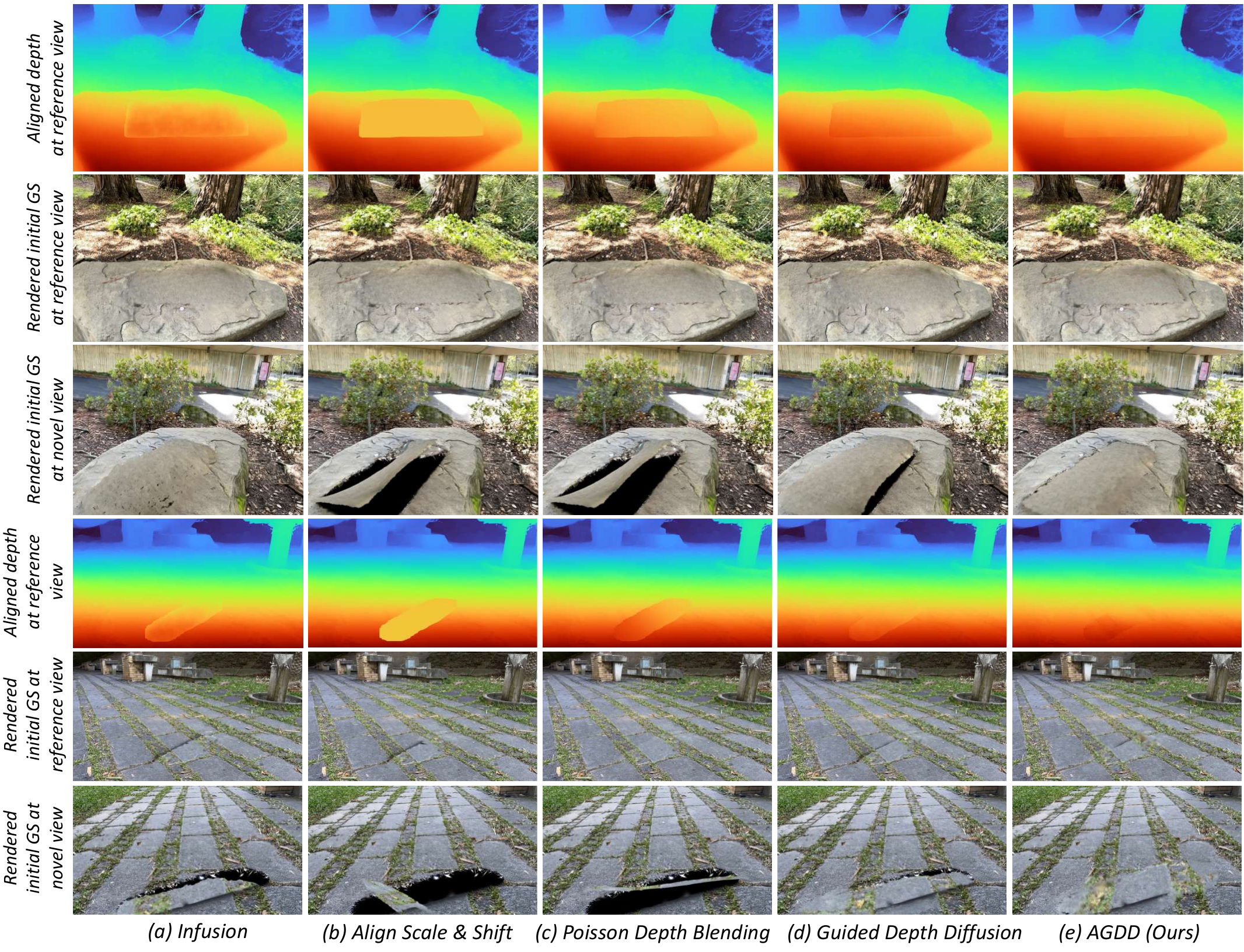}
    \vspace{-6mm}
    \caption{\textbf{Compared to other depth completion methods.} The depth completion model in Infusion~\cite{liu2024infusion} (a) performs better at depth alignment compared to traditional methods (b) and (c), but it produces noisy depth in unseen regions. Similarly, (d) Guided Depth Diffusion~\cite{yu2024wonderworld} struggles to achieve precise alignment, as the background regions amplify the loss, leading to misalignment. In contrast, (e) Our AGDD effectively addresses these issues.} %
    \label{fig:ablation_depth_alignment2}
\end{figure}

\vspace{3pt}
\noindent {\bf Effect of reference view initial Gaussians alignment.}
\cref{tab:ablation} and ~\cref{fig:ablation_depth_alignment2} show that our depth-aware 3DGS initialization accurately estimates aligned depth while maintaining geometric consistency in the inpainted regions. Compared to random initialization, our method produces more structurally coherent results, particularly in areas with significant depth variations. This is especially evident in scenes where the inpainted geometry needs to blend seamlessly with the existing scene structure.




\section{Conclusion}
\label{sec:conclusion}

We presented AuraFusion360, a novel reference-based 360° inpainting method for 3D scenes in unbounded environments. Our approach effectively addresses the challenges of object removal and hole filling in complex 3D scenes. Key contributions include leveraging multi-view information through improved unseen mask generation, integrating reference-guided 3D inpainting with diffusion priors, and introducing the 360-USID dataset for comprehensive evaluation.
Experimental results demonstrate AuraFusion360's superior performance over existing methods, particularly in complex geometries and large view variations. While this work represents a significant advancement in 3D scene editing, future work will focus on computational efficiency, dynamic scenes, and language-guided editing capabilities.

\newpage
\paragraph{Acknowledgements.}
This work was supported by NVIDIA Taiwan AI Research \& Development Center (TRDC).
This research was funded by the National Science and Technology Council, Taiwan, under Grants NSTC 112-2222-E-A49-004-MY2 and 113-2628-E-A49-023-. Yu-Lun Liu acknowledges the Yushan Young Fellow Program by the MOE in Taiwan.

{\small
\bibliographystyle{ieeenat_fullname}
\bibliography{11_references}
}

\ifarxiv \clearpage \appendix \label{sec:appendix_section}

\section{Overview}
This supplementary material provides additional details and results to support the main manuscript. We first describe the training process for masked Gaussians and object removal in Section~\ref{sec:masked-gs}, followed by an explanation of depth warping for bounding box generation in SAM2~\cite{ravi2024sam2} and its role in identifying unseen region contours in Section~\ref{sec:depth_warping}. Next, we present ablations on different depth inpainting methods in Section~\ref{sec:mad} and a comparison of captured and inpainted references in Section~\ref{sec:ref_require}. We then outline the experimental setup in Section~\ref{sec:compared_method} and discuss the limitations of our approach in Section~\ref{sec:limitations}. Finally, we provide additional visual comparisons in ~\cref{fig:visual_add_ours} for the 360-UISD dataset and in ~\cref{fig:visual_add_360} for the other collected 360 dataset~\cite{barron2022mip}.


\section{Training Masked GS for Object Removal}
\label{sec:masked-gs}
During the training of masked Gaussians, we use 2DGS~\cite{huang20242d} as our codebase and introduce a masked attribute, ranging between 0 and 1, for each Gaussian. The L1 loss is computed between the object mask obtained via SAM2~\cite{ravi2024sam2} and the rasterized object mask for each training view. Additionally, we incorporate the Grouping Loss proposed by Gaussian Grouping~\cite{ye2023gaussian}, ensuring that neighboring Gaussians have similar masked attributes. This ensures that our Gaussian model retains accurate object mask information and is capable of rendering precise object masks for subsequent applications.

Thanks to the explicit nature of Gaussian Splatting, we can directly remove Gaussians with a masked attribute greater than a threshold $\tau$ during the removal stage, effectively achieving object removal. In our implementation, $\tau$ is set to 0.6.

\begin{figure}[t]
    \centering
    \includegraphics[width=1\linewidth]{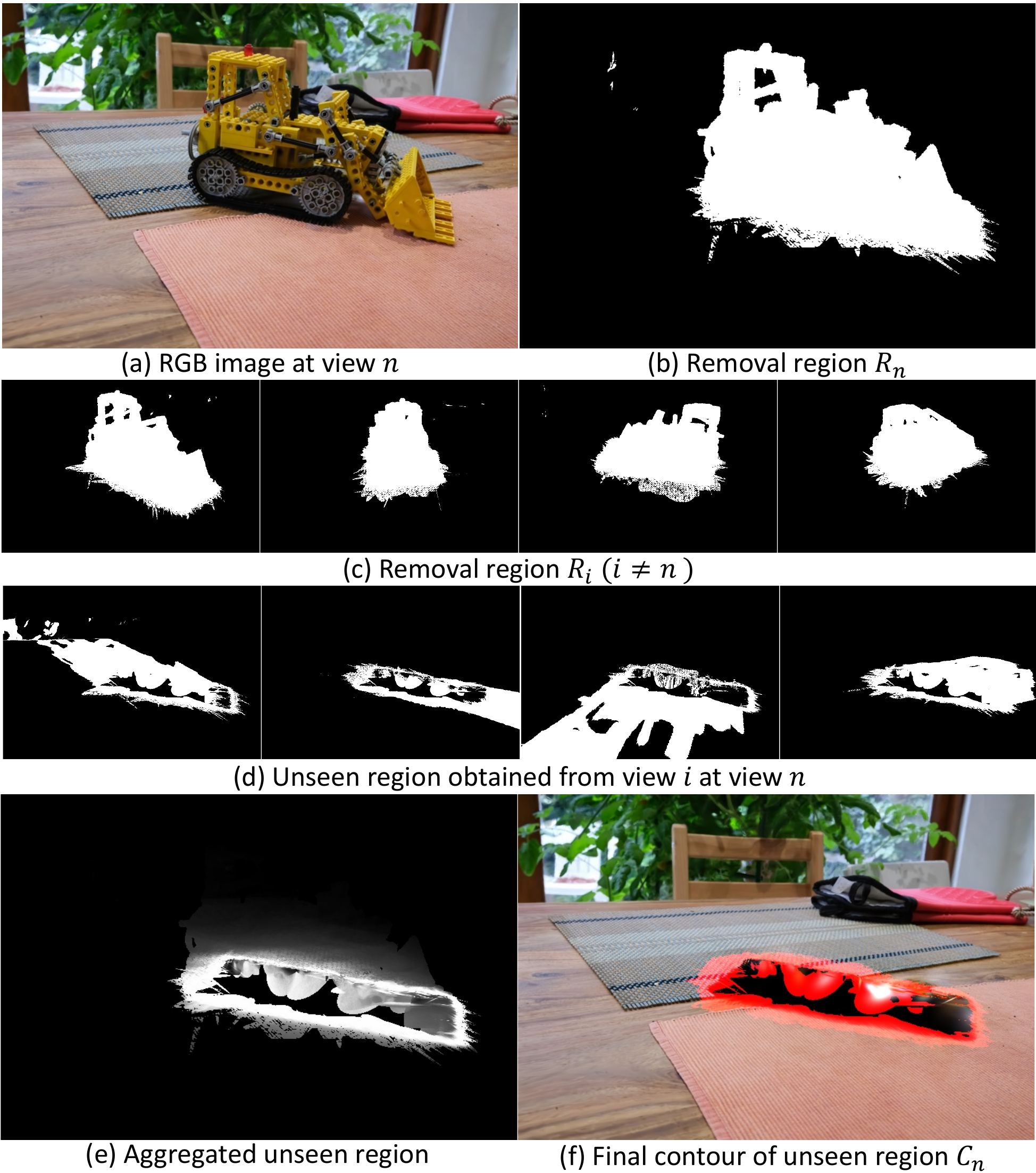}
    \caption{\textbf{Intermediate Results of Depth Warping for Unseen Region Detection.} This figure illustrates the intermediate results generated during the depth warping process. (a) and (b) show the RGB image and the corresponding removal region at view $n$, respectively. (c) displays the removal regions obtained from view $i$ ($i \neq n$). (d) shows the unseen region obtained from view $i$ through backward traversal. The intersections are concentrated near the unseen region. Note that the pixels within the unseen region, but with a value of zero, are due to the absence of Gaussians in that area, preventing depth rendering and thus making it impossible to establish pixel correspondences between view $n$ and view $i$. (e) presents the aggregation of all unseen regions obtained from view $i$ at view $n$. A threshold is applied to this result, and it is then intersected with the removal region at view $n$ to obtain the final result in (f).}

    \label{fig:unseen_region}
\end{figure}

\section{Depth Warping for Unseen Contours}

\label{sec:depth_warping}

Following Sec. \textcolor{cvprblue}{3.2} and Fig. \textcolor{cvprblue}{4} of the main paper, we explain in detail how depth warping allows us to identify the contours of the unseen region, as illustrated in ~\cref{fig:unseen_region}. Without loss of generality, to find the unseen region contour at view \(n\), and for each pair of views \(n\) and \(i\), we first compute the removal region for view \(i\) by identifying pixels that differ between the rendered depth and the incomplete depth of view \(i\) rather than using object masks. This approach better captures geometric changes and prevents misalignment artifacts, leading to improved SAM2\cite{ravi2024sam2} prompts and more precise unseen masks (\cref{fig:unseen_counter_ablation}). 

Next, we establish pixel correspondences between view \(n\) and view \(i\) using the incomplete depth of view \(n\). The removal region of view \(i\) is then backward-traversed to view \(n\) based on these correspondences. During this backward traversal, it is important to note that pixels outside the unseen region in view \(i\) will correspond to the background areas in view \(n\), while pixels belonging to the unseen region remain in the unseen region. By aggregating contributions from all views \(i\) (\(i \neq n\)), we project non-unseen regions from each view \(i\) into different areas of view \(n\), while consolidating the unseen regions. This allows us to identify the contours of the unseen region in view \(n\). These contours can then be used as the bounding box prompt for SAM2, resulting in a more accurate unseen mask. 


\begin{figure}[t]
    \centering
    \includegraphics[width=1\linewidth]{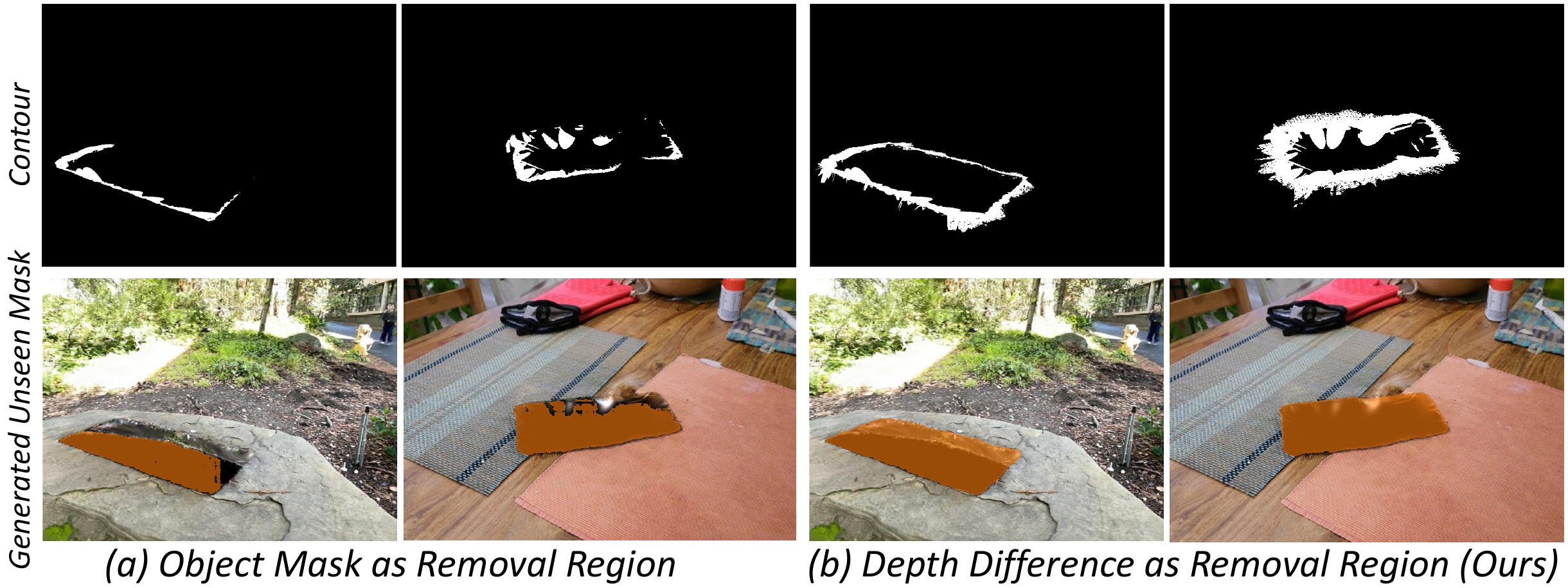}
    \caption{\textbf{Ablation Study on Removal Region Definition.} Comparison of (a) object masks vs. (b) depth difference for defining removal regions. Object masks fail to capture geometric changes, leading to less accurate unseen masks. Depth difference better preserves scene structure, improving SAM2 prompts and unseen region segmentation.}
    \label{fig:unseen_counter_ablation}
\end{figure}


\section{Comparison of Depth Completion Methods}
\label{sec:mad}
In addition to Fig. \textcolor{cvprblue}{11} of the main paper, we compare scale–shift alignment, LaMa~\cite{lama}, InFusion~\cite{liu2024infusion}, GDD~\cite{yu2024wonderworld}, and AGDD for depth completion. As shown in ~\cref{tab:mad_quan}, we evaluate the mean absolute difference (MAD) in object mask areas in 30 test views, using pseudo-GT depth from a 2DGS trained on 200 removal images, as mentioned in Sec. \textcolor{cvprblue}{4}.
Aligning scale-shift misaligns boundaries in 360° scenes, while LaMa provides reasonable depth completion but does not fully resolve alignment issues. AGDD achieves the lowest MAD and better handles complex geometry.

\begin{table}[h]
\centering
\tiny
\caption{MAD values for different depth completion methods.} 
\label{tab:mad_quan}  
\resizebox{\columnwidth}{!} {
\begin{tabular}{p{4cm}|c} 
\toprule
Depth completion method & MAD $\downarrow$ \\
\midrule
Scale-shift align & 0.063 \\
LaMa depth inpainting & 0.077 \\ 
InFuion & 0.047 \\
GDD & 0.065\\ 
AGDD & \textbf{0.045} \\
\bottomrule
\end{tabular}
}
\vspace{-1.3pc}
\end{table}

\section{Reference Images in Real-World Use}
\label{sec:ref_require}
Our 360-USID dataset provides real-world captured reference images. However, this does not mean that our method requires extra input. In practical scenarios, reference images can be captured post-removal for real-world use. We also ensure a fair evaluation by avoiding hallucinated textures, even if the inpainting is consistent. Additionally, reference guidance helps reduce multi-view inconsistency with minimal extra input.
As shown in ~\cref{tab:ref_inpain_quan}, while LaMa-based references slightly degrade the results, they still outperform other reference-based methods, such as GScream. Even when using an inpainted image as a reference, our approach still achieves good results.

\begin{table}[h]
\centering
\footnotesize
\caption{Comparison of Captured and Inpainted Reference.}
\label{tab:ref_inpain_quan}  
\resizebox{\columnwidth}{!} {
\begin{tabular}{l|cccc}
\toprule
Reference method & PSNR $\uparrow$ & SSIM $\uparrow$ & LPIPS $\downarrow$ & FID $\downarrow$ \\
\midrule
GScream & 14.758 & 0.955 & 0.514 & 152.295 \\
LaMa-reference & 17.102 & 0.960 & 0.407 & 69.874 \\ 
Captured-reference &  \textbf{17.661} & \textbf{0.961}  & \textbf{0.388} & \textbf{62.173} \\
\bottomrule
\end{tabular}
}
\vspace{-1pc}
\end{table}

\section{Experimetal Setup}
\label{sec:compared_method}

\subsection{LeftRefill~\cite{cao2024leftrefill}}
We use the same reference image as in our method, along with the rendered object masks of each novel testing view generated by our masked Gaussians, as input to LeftRefill and directly perform reference-based inpainting on each testing novel view.

\subsection{2DGS~\cite{huang20242d} + LaMa~\cite{lama}}
We provide the same reference image and training view object masks as in our method and use LaMa~\cite{lama} to obtain per-frame inpainting results for each training view to train the 2DGS.

\subsection{2DGS~\cite{huang20242d} + LeftRefill~\cite{cao2024leftrefill}}
We provide the same reference image and training view object masks as in our method and use LeftRefill to obtain per-frame inpainting results for each training view to train the 2DGS.

\begin{figure}[t]
    \centering
    \includegraphics[width=1\linewidth]{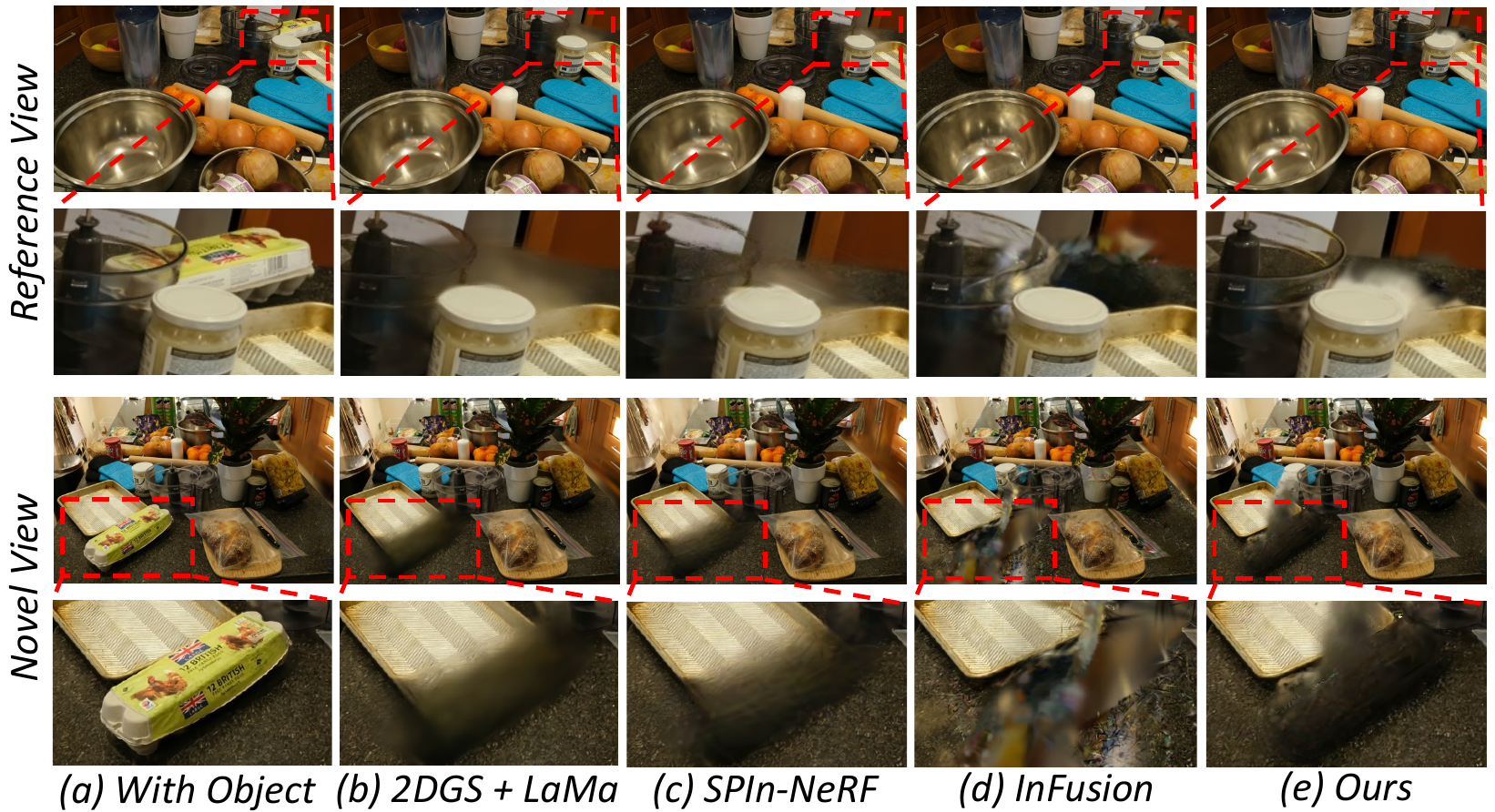}
    \caption{\textbf{Failure Cases.} The figure illustrates failure cases of inpainting results. These examples highlight the challenges of 3D inpainting when significant occlusions are present near the regions requiring inpainting. For instance, (b) and (c) demonstrate difficulties in achieving satisfactory guided inpainted RGB images in the training views, while (d) and (e) show errors resulting from incorrect pixel unprojections. These observations indicate that this issue is not effectively addressed by any of the compared methods, suggesting a potential avenue for further exploration and improvement.}


    \label{fig:fail_case}
\end{figure}

\subsection{SPIn-NeRF~\cite{spinnerf}}
The original SPIn-NeRF~\cite{spinnerf} codebase is designed for forward-facing scenes; however, we adapt it for comparison on 360° scenes by implementing its approach on 2DGS~\cite{huang20242d}. We first obtain the depth for each training view by training a 2DGS model. Next, we generate inpainted RGB and depth maps using LaMa~\cite{lama}, which are then used to train the inpainted 2DGS model. During training, we follow SPIn-NeRF's methodology by incorporating patch-based RGB-LPIPS loss and using the Pearson correlation coefficient to compute a scale- and shift-invariant depth loss.

\subsection{Gscream~\cite{wang2024gscream}}
We follow the original GScream~\cite{wang2024gscream} pipeline as a baseline for comparison. We provide the same reference image and training view object masks as our method to ensure consistency. Following their pipeline, we use Marigold~\cite{ke2023repurposing} to generate estimated depths for all training images, meeting GScream's input data requirements.

\begin{figure*}[t]
    \centering
    \includegraphics[width=1\linewidth]{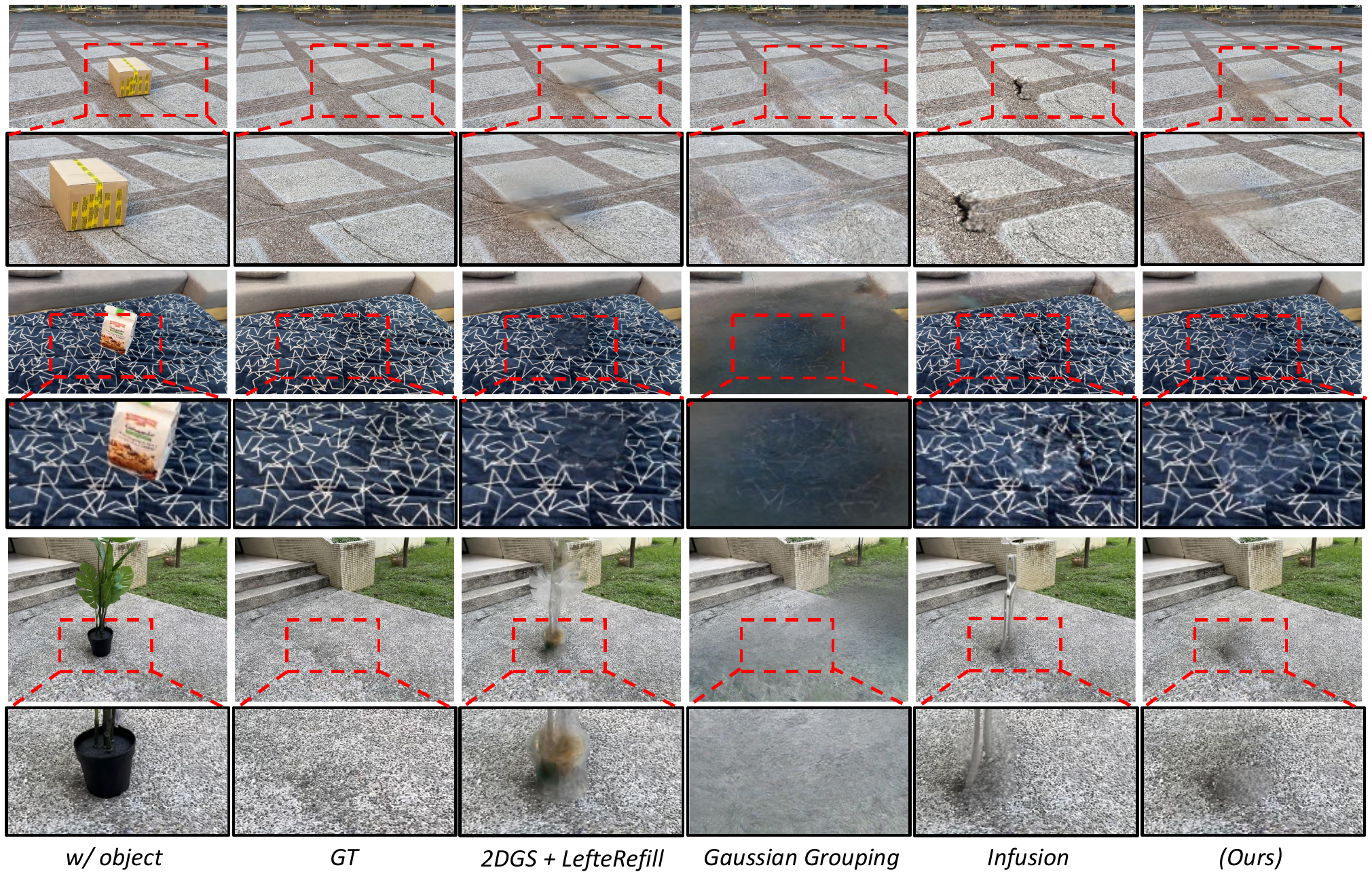}
    \caption{\textbf{Visual Comparison on our 360-USID dataset.} }
    \label{fig:visual_add_ours}
\end{figure*}

\subsection{Gaussian Grouping~\cite{wang2024gscream}}

We utilize the original Gaussian Grouping~\cite{ye2023gaussian} codebase as a baseline for comparison. First, it generates segmentation IDs, from which we select the IDs corresponding to objects that require inpainting. These selected IDs are then used in the removal process. Following the original workflow, the unseen regions are identified, subsequently inpainted, and used for their fine-tuning process.

Notably, after removing objects from the scene, Gaussian Grouping relies on TrackingAnything-DEVA~\cite{cheng2023tracking} to identify unseen regions requiring further inpainting through the "black blurry hole" prompt. However, DEVA occasionally fails to accurately identify unseen regions in certain scenes, leading to incorrect inpainting and suboptimal results. Additionally, in some scenes, such as the \textit{bonsai} scene from the Mip-NeRF-360~\cite{barron2022mip} dataset and the \textit{plant} scene from the 360-UISD dataset, the object tracker misidentifies objects, resulting in incorrect object removal and further degrading the inpainting quality.

\subsection{InFusion~\cite{liu2024infusion}}
We use the original InFusion~\cite{liu2024infusion} codebase as a baseline for comparison. We provide the same reference image used in our method as the input RGB for its depth completion model. This reference image is also used in its fine-tuning process.




\section{Limitations}
\label{sec:limitations}
Our method successfully addresses complex, unbounded 360° scene inpainting. However, rendering the unprojected initial Gaussians and applying SDEdit~\cite{meng2022sdedit} to enhance the guided inpainted RGB images can be time-consuming, particularly for high-resolution or large-scale scenes, which poses challenges for real-time applications. Furthermore, our analysis~\cref{fig:fail_case} shows that the method may produce incorrect pixel unprojections in cases with significant occlusions near the object requiring inpainting, resulting in floaters in the final inpainted outputs. This limitation is similarly observed across all compared methods, underscoring a valuable direction for future research and improvement.


\begin{figure*}[t]
    \centering
    \includegraphics[width=1\linewidth]{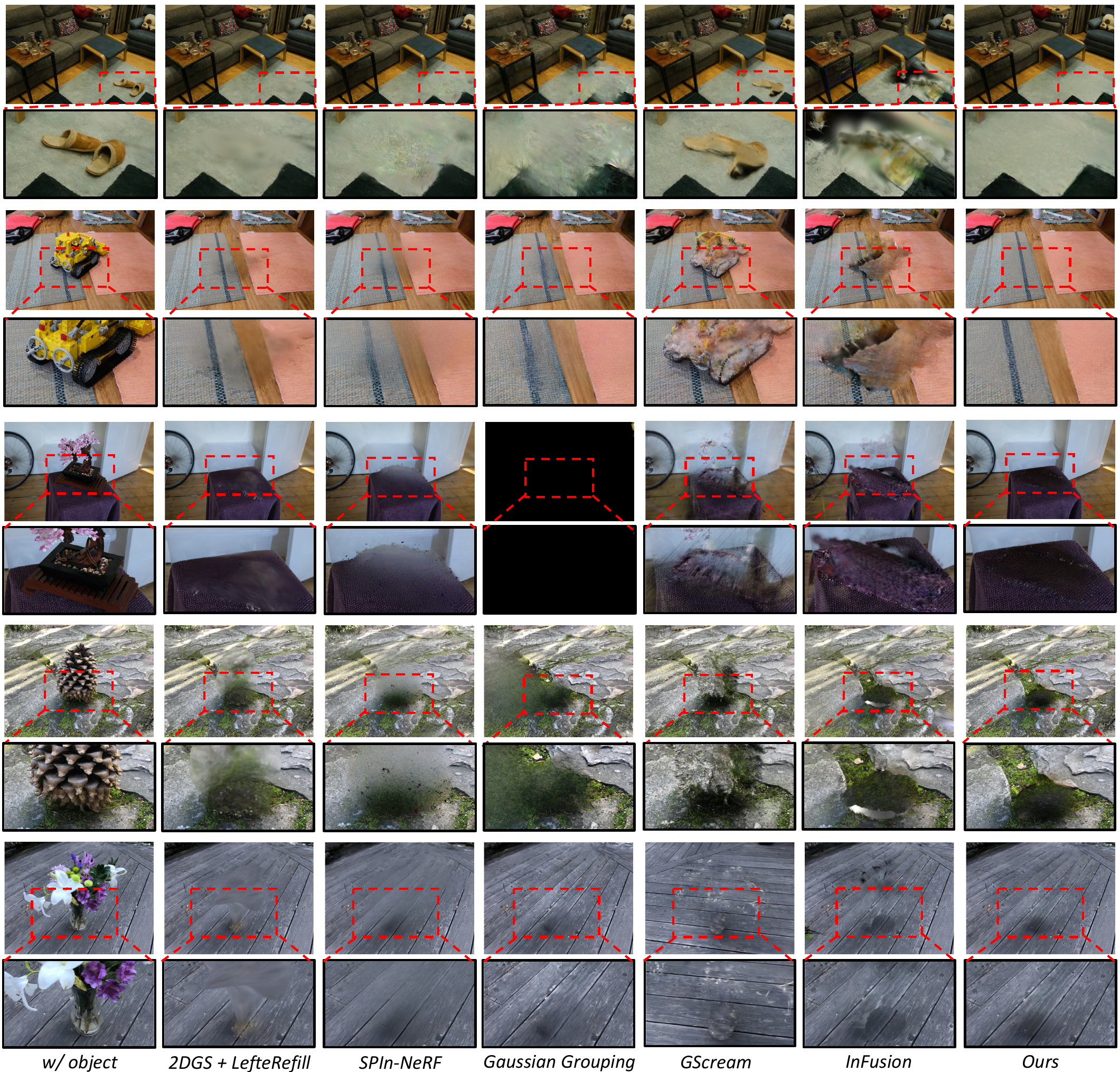}
    \caption{\textbf{Visual Comparison on Other-360 dataset.} }
    \label{fig:visual_add_360}
\end{figure*}

 \fi

\end{document}